# multimodars: A Rust-powered toolkit for multi-modality cardiac image fusion and registration

Anselm W. Stark, MD[a,b,∗], Marc Ilic, MSc[a,b], Ali Mokthari, MSc[a,b], Pooya Mohammadi Kazaj, MSc[a,b], Christoph Gräni, MD, Ph.D.[a], Isaac Shiri, Ph.D.[a]

[a]Department of Cardiology, Inselspital, Bern University Hospital, University of Bern, Switzerland
[b] Graduate School for Cellular and Biomedical Sciences, University of Bern, Bern, Switzerland

## 1. Summary

Coronary artery anomalies (CAAs) and coronary artery disease (CAD) require precise morphological and functional assessment for diagnosis and treatment planning. Cardiac computed tomography angiography (CCTA) provides comprehensive 3D coronary anatomy but lacks the sub-millimetre resolution and dynamic tissue detail available from intravascular imaging (intravascular ultrasound [IVUS], optical coherence tomography [OCT]). Developed initially to quantify dynamic lumen changes in CAAs, multimodars is a general-purpose toolkit that registers high-resolution intravascular pullbacks to CCTA-derived centerlines, producing locally enhanced fusion 3D vessel representations for visualization, geometric analysis, and patient-specific modelling. The toolkit implements four alignment paradigms (full, double-pair, single-pair, single) to compare pullbacks acquired under different haemodynamic states (e.g., rest vs. pharmacologic stress) or at different timepoints (e.g., pre- vs. post-stenting). multimodars targets deterministic, reproducible multimodal fusion for CAA research and general CAD applications Stark et al. (2025b,a) (see Figure 1).

∗Corresponding Author: Anselm Stark, MD
PhD student
Department of Cardiology
University Hospital Bern
Freiburgstrasse
CH - 3010 Bern, Switzerland
Email: anselm.stark@insel.ch

## 2. Statement of need

Combining complementary imaging modalities is critical to build reliable 3D coronary models: intravascular imaging gives sub-millimetre resolution but limited whole-vessel context, while CCTA supplies 3D geometry but suffers from limited spatial resolution and artefacts (e.g., blooming). Prior work demonstrated intravascular/CCTA fusion van der Giessen et al. (2010); Kilic et al. (2020); Wu et al. (2020); Bourantas et al. (2013), yet no open, flexible toolkit is tailored for multi-state analysis (rest/stress, pre-/post-stenting) while offering deterministic behaviour, high performance, and easy pipeline integration. multimodars addresses this gap with deterministic alignment algorithms, a compact NumPy-centred data model, and an optimised Rust backend suitable for scalable, reproducible experiments. The package accepts CSV/NumPy inputs including data formats produced by the AIVUS-CAA software Stark et al. (2025c).

## 3. Features

multimodars combines a Rust core with a compact typed Python data model (PyContourPoint, PyContour, PyGeometry, PyGeometryPair) that maps losslessly to (N,4) NumPy arrays (*frame_id, x, y, z*) (see Figure 2). Key capabilities: round-trip CSV/NumPy I/O, optional OBJ export with deformation mapping and UV coordinates, centroiding, area and ellipticity metrics, smoothing, reordering, rigid transforms, stenosis summaries, and mesh export. Processing modes return either all four geometry pairs (full), pulsatile deformation in two haemodynamic states (double-pair), any pair of states (single-pair), or intra-pullback alignment only (single). Users tune down sampling and angular resolution to trade accuracy for throughput; computational hotspots are parallelised.

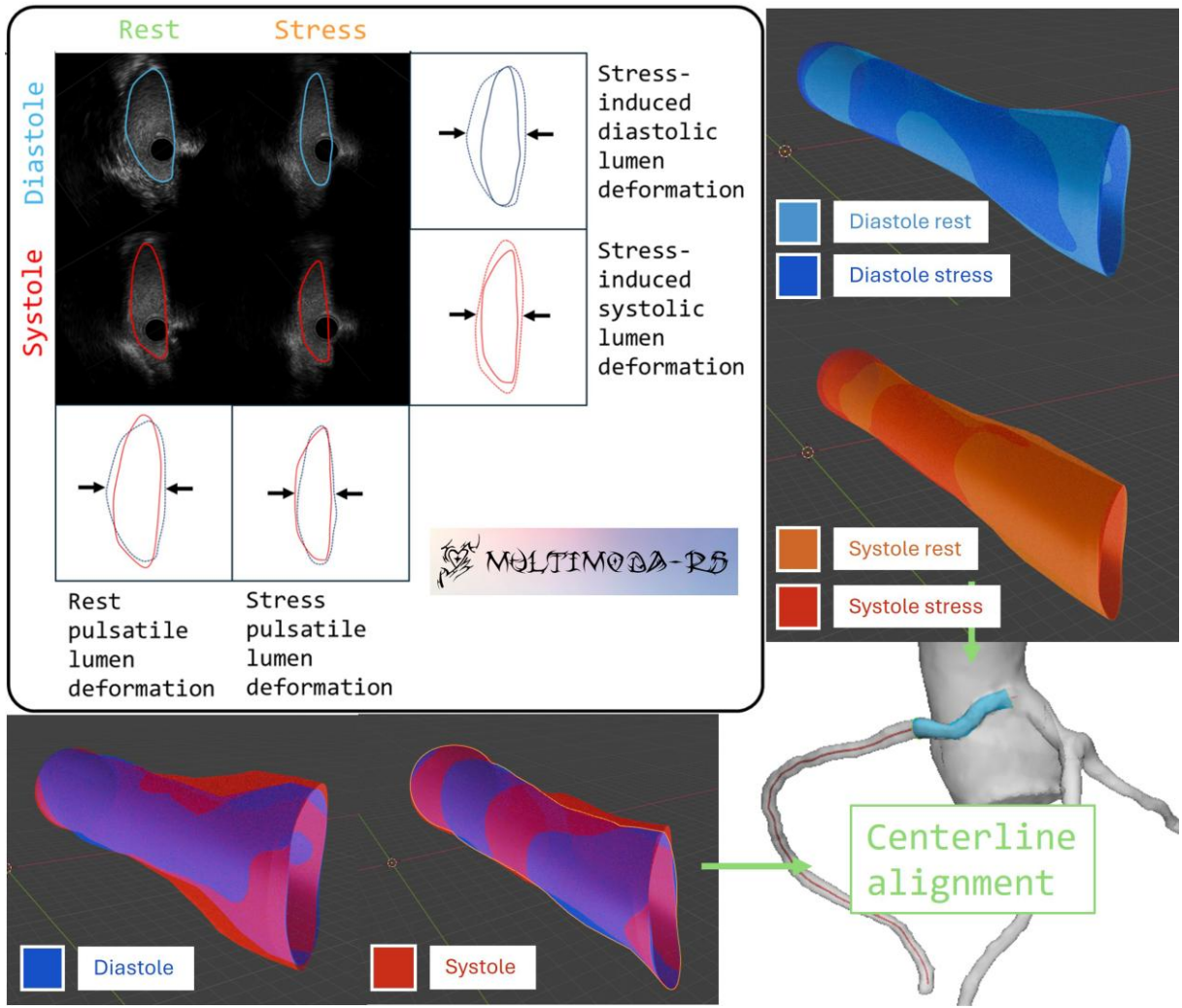

**Figure 1:** Illustration of multimodars processing modes and their clinical use. The 'full' mode returns four geometry pairs to analyze rest and stress haemodynamics (rest pulsatile deformation; stress pulsatile deformation; stress-induced diastolic deformation; stress-induced systolic deformation). The 'double-pair' mode returns pulsatile deformation in rest and stress. 'Single-pair' compares any two states (e.g., pre-/post-stent). 'Single' aligns frames within one pullback

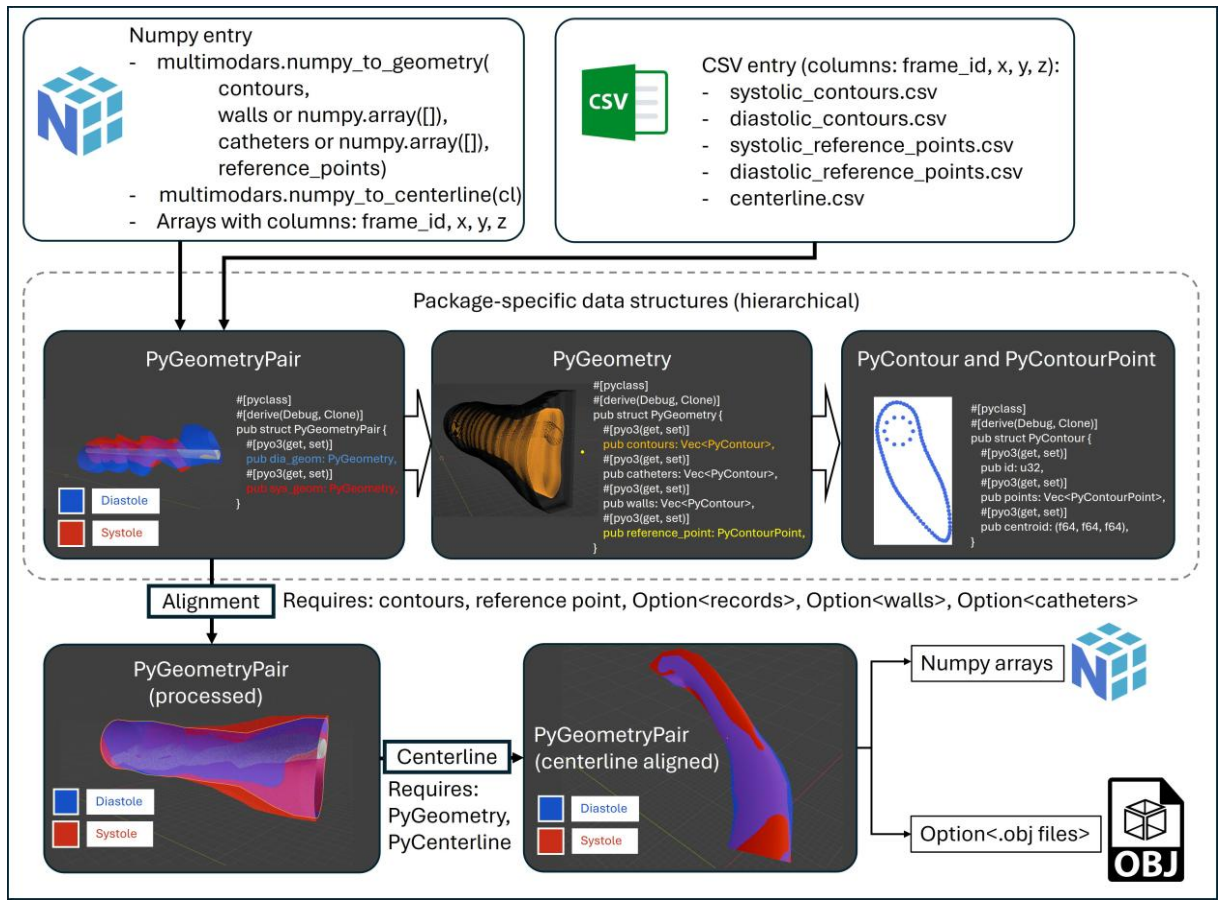

**Figure 2:** Internal data model and (N,4) NumPy mapping used by multimodars (*frame_id, x, y, z*).

## 4. Algorithms

Alignment is a two-stage pipeline producing spatially and rotationally consistent mappings both within pullbacks (intra-pullback) and between pullbacks (inter-pullback).

- **Intra-pullback:** The proximal frame is the rotational reference. Sequentially, each proximal→distal neighbour is aligned by centroid translation and a rotation search minimizing a point-set distance derived from directed Hausdorff distances. Rotation employs a multiscale angular search (coarse → fine, e.g., 1° → 0.1° → 0.01°) with cumulative rotation propagation to preserve vessel torsion (See Figure 3). Naive brute-force complexity scales as $O(n \times \frac{R}{n} \times m^2)$ (n = frames, m = points per contour, R = angular range, S = step size). By fixing contour size (downsampling) and reducing the angular search via multiscale refinement, the pipeline attains effective complexity $O(n \times (R + c) \times m^2)$ for small S, making runtime less sensitive to step granularity while preserving alignment accuracy.

- **Inter-pullback and CCTA fusion:** Inter-pullback alignment harmonizes distal centroids, averages slice spacing to align z-coordinates, and applies a rigid rotation to minimize mean directed distances across corresponding frames; ellipticity-weighted similarity prioritizes non-round stenotic slices. For CCTA, multimodars implements a three-point anatomical registration and a manual alignment mode for ambiguous anatomies: centerlines are resampled to contour spacing, centroids are translated to matched points, normals are aligned by cross-product computations, and an optional interpolated UV-mapped mesh is produced for visualization and downstream modelling. Clinical impact in CAA and beyond: multimodars was motivated by quantifying dynamic lumen deformation in CAAs, where rest/stress and pulsatile comparisons are diagnostically relevant. Deterministic, high-resolution fusion enables quantitative assessment of stress-induced deformation and supports planning and patient-specific haemodynamic modelling. Identical methods support longitudinal CAD analyses (e.g., pre-/post-stent) and broader cardiovascular research.

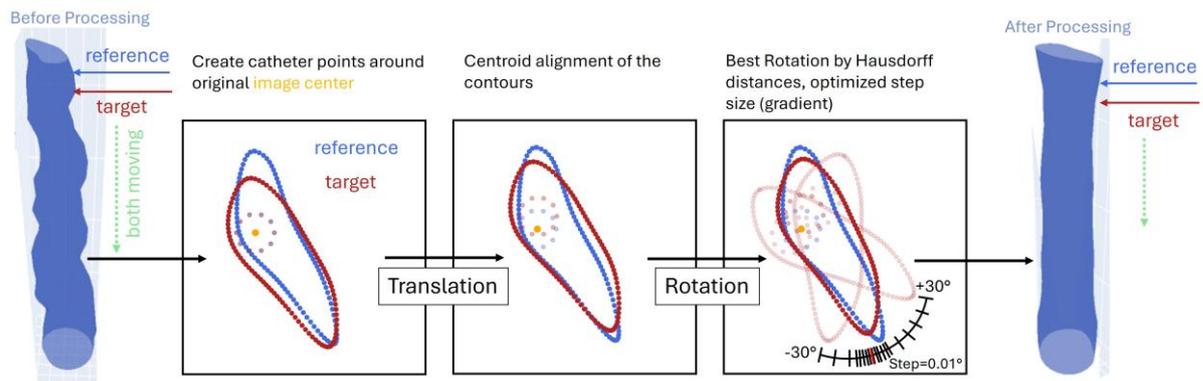

**Figure 3:** Multiscale intra-pullback alignment workflow (coarse-to-fine angular search and centroid propagation).

## 5. Performance and parallelisation

Rust (Rayon) provides hierarchical data parallelism and SIMD-enabled coordinate transforms. Point rotations and nearest-neighbour searches parallelize across cores; independent pullbacks and frames are processed concurrently when dependencies allow. Typical production workflows downsample contours to 200–500 points/frame to balance sub-pixel accuracy and compute.

Empirical performance (representative benchmark on consumer hardware): on a 16-core CPU, an OCT pullback with 280 frames, a rotation search range of ±3°, and a final angular accuracy of 0.01° saw alignment time reduced from ≈150s (multi-threaded, bruteforce search) to ≈18s with the optimized, parallel coarse search (mean frame rotation difference −0.01°). These numbers illustrate practical speedups available by combining downsampling, multiscale search and parallelism; exact timings depend on hardware, parameters and input size.

## 6. Implementation, reproducibility and usage

The core is implemented in Rust and exposed to Python via PyO3; packaging uses maturin and the package is available on PyPI. The NumPy-centric API maps directly to (N,4) arrays; the project ships example notebooks, sample data (including AIVUS-CAA Stark et al. (2025c)) and CI tests. Documentation and tutorials are hosted on ReadTheDocs (https://multimoda-rs.readthedocs.io/en/latest/) to support reproducible workflows. Users can tune down sampling, catheter reconstruction, and angular resolution to balance precision and throughput.

## 7. Acknowledgements

None

# References

Stark, A.W., Bigler, M.R., Räber, L., and Gräni, C., True Pulsatile Lumen Visualization in Coronary Artery Anomalies Using Controlled Transducer Pullback and Automated IVUS Segmentation, Case Reports, 30(22):104741, 2025. doi:10.1016/j.jaccas.2025.104741.

Stark, A.W., Bigler, M.R., Kakizaki, R., Räber, L., and Gräni, C., IVUS Is Superior to OCT for the Visualization of Dynamic Anatomic Vessel Lumen Changes in Anomalous Coronary Artery, Catheterization and Cardiovascular Interventions, 2025. doi:10.1002/ccd.31660.

van der Giessen, A.G., Schaap, M., Gijsen, F.J.H., Groen, H.C., van Walsum, T., Mollet, N.R., Dijkstra, J., van de Vosse, F.N., Niessen, W.J., de Feyter, P.J., et al., 3D fusion of intravascular ultrasound and coronary computed tomography for in-vivo wall shear stress analysis: a feasibility study, The International Journal of Cardiovascular Imaging, 26(7):781–796, 2010. doi:10.1007/s10554-009-9546-y.

Kilic, Y., Safi, H., Bajaj, R., Serruys, P.W., Kitslaar, P., Ramasamy, A., Tufaro, V., Onuma, Y., Mathur, A., Torii, R., et al., The evolution of data fusion methodologies developed to reconstruct coronary artery geometry from intravascular imaging and coronary angiography data: a comprehensive review, Frontiers in Cardiovascular Medicine, 7:33, 2020. doi:10.3389/fcvm.2020.00033.

Wu, W., Samant, S., De Zwart, G., Zhao, S., Khan, B., Ahmad, M., Bologna, M., Watanabe, Y., Murasato, Y., Burzotta, F., et al., 3D reconstruction of coronary artery bifurcations from coronary angiography and optical coherence tomography: feasibility, validation, and reproducibility, Scientific Reports, 10(1):18049, 2020. doi:10.1038/s41598-020-74264-w.

Bourantas, C.V., Papafaklis, M.I., Athanasiou, L., Kalatzis, F.G., Naka, K.K., Siogkas, P.K., Takahashi, S., Saito, S., Fotiadis, D.I., Feldman, C.L., et al., A new methodology for accurate 3-dimensional coronary artery reconstruction using routine intravascular ultrasound and angiographic data: implications for widespread assessment of endothelial shear stress in humans, EuroIntervention, 9(5):582–593, 2013. doi:10.4244/EIJV9I5A94.

Stark, A.W., Mohammadi Kazaj, P., Balzer, S., Ilic, M., Bergamin, M., Kakizak i, R., Giannopoulos, A., Haeberlin, A., Räber, L., Shiri, I., et al., Automated IntraVascular UltraSound Image Processing and Quantification of Coronary ArteryZ Anomalies: The AIVUS-CAA software, *Computer Methods and Programs in Biomedicine*, 272, 109065 (2025). https://doi.org/10.1016/j.cmpb.2025.109065

# Welcome to multimodars documentation!

*multimodars: A Rust-powered cardiac multi-image modality fusion package.*

This package aims to combine different cardiac imaging modalities to combined 3D models. A particular focus is on the fusion of intravascular imaging with computed coronary tomography angiography (CCTA).

> **❶ Warning**
>
> Not intended for clinical use.

## Table of Contents

## License

This package is covered by the open source MIT License.

## Developers

- Anselm Stark[1,2]
- Marc Ilic[1,2]
- Ali Mokthari[1,2]
- Pooya Mohammadi Kazaj[1,2]
- Isaac Shiri[1]

[1]Department of Cardiology, Inselspital, Bern University Hospital, University of Bern, Switzerland
[2]Graduate School for Cellular and Biomedical Sciences, University of Bern, Bern, Switzerland

## Contributing

We'd welcome your contributions to multimodars. Please read the contributing guidelines on how to contribute to `multimodars`.

## Indices and tables

- Index
- Search Page

# Installation

There are two ways you can use pyradiomics: 1. Install via pip 2. Install from source

## 1. Install via pip

Pre-built binaries are available on PyPI for installation via pip. For the python versions mentioned below, wheels are automatically generated for each release of `multimodars`, allowing you to install multimodars without having to compile anything.

- Ensure that you have `python` installed on your machine, version 3.12 or higher (64-bits).
- Install multimodars:

```
python -m pip install multimodars
```

## 2. Install from source

multimodars can be installed from source with the following steps.

- Clone the repository and install Rust and Maturin:

```
# Install rust in case you don't have it on your system
curl --proto '=https' --tlsv1.2 -sSf https://sh.rustup.rs | sh

git clone https://github.com/yungselm/multimoda-rs.git
python -m venv .venv
source .venv/bin/activate
pip install maturin
. "$HOME/.cargo/env" # Set rust env
maturin develop
```

> **ℹ Note**
>
> In case you get the following error:
>
> ```
> 💥 maturin failed
> Caused by: rustc, the rust compiler, is not installed or not in PATH.
> This package requires Rust and Cargo to compile extensions. Install it
> through the system's package manager or via https://rustup.rs/.
> ```
>
> execute the following commands:

```
unset -v VIRTUAL_ENV
maturin develop
```

# Tutorial

This step-by-step tutorial demonstrates how to:

- Run the workflow from csv files
- Run the workflow by building geometries from numpy arrays
- Finetuning of alignment algorithms
- Alignment with a centerline
- Saving everything as .obj files
- Utility functions to link to numpy
- Reordering algorithm
- Class methods

## 1. Workflow csv files

After pip installing or locally building the package, install it in the familiar way.

```python
import multimodars as mm
```

To run the whole workflow from .csv files the following requirements have to be met. Files should be named `diastolic_contours.csv`, `systolic_contours.csv`, `diastolic_reference_points.csv` and `systolic_reference_points.csv` depending on the required analysis. Every file should be structured in the following way (no headers):

| ... | ... | ... | ... |
|-----|-----|-----|-----|
| 771 | 2.4862 | 6.7096 | 24.5370 |
| 771 | 2.5118 | 6.7017 | 24.5370 |
| 771 | 2.5370 | 6.6936 | 24.5370 |
| ... | ... | ... | ... |

To acquire meaningful measurement data, the coordinates should be provided in mm or another SI unit instead of pixel values. Optionally a record file can be provided *combined_sorted_manual.csv*, which should have the following structure, here the first column should contain the desired frame order and "measurement_1" represent the thickness of the wall between aorta and coronary and "measurement_2" for the thickness between pulmonary artery and coronary (position just for demonstration). This is based on the output of the AIVUS-CAA software:

| frame | (position) | phase | measurement_1 | measurement_2 |
|-------|-----------|-------|---------------|---------------|
| 18 | 23.99 | D | | |
| 37 | 22.79 | D | | 2.35 |

| 212 | 21.59 | D | 1.38 | 2.34 |
|-----|-------|---|------|------|
| 94  | 20.39 | D | 1.38 | 2.11 |
| ... | ...   | ... | ... | ... |
| 47  | 18.78 | S |      |      |

This simplifies the workflow, by just providing a directory to automatically process:

```
rest, stress, dia, sys, (rest_logs, stress_logs, dia_logs, sys_logs) = mm.from_file(
    mode="full",
    rest_input_path="rest_csv_files",
    stress_input_path="stress_csv_files",
    rest_output_path="output/rest",
    stress_output_path="output/stress",
)
```

However the preferred more flexible way is from numpy arrays.

## 2. Workflow numpy arrays and Finetuning

Here the geometry can be directly build from arrays with the same structure as before:

| ...  | ...    | ...    | ...     |
|------|--------|--------|---------|
| 771  | 2.4862 | 6.7096 | 24.5370 |
| 771  | 2.5118 | 6.7017 | 24.5370 |
| 771  | 2.5370 | 6.6936 | 24.5370 |
| ...  | ...    | ...    | ...     |

catheter and walls are optional. However it is not recommended to provide the catheter points directly, but rather the image center (in mm), radius of the catheter (e.g. 0.5mm for IVUS) and number of points to represent the catheter. If no walls are provided a default wall with 1mm offset is created.

```
prestent = mm.numpy_to_geometry(
    contours_arr=contours,
    catheters_arr=np.array([]),
    walls_arr=np.array([]),
    reference_arr=references,
)

poststent = mm.numpy_to_geometry(
    contours_arr=contours,
    catheters_arr=np.array([]),
    walls_arr=np.array([]),
    reference_arr=references,
)

pair, logs = mm.from_array(
    mode="singlepair",
    geometry_dia=prestent,
    geometry_sys=poststent,
    step_rotation_deg=0.1,
    range_rotation_deg=30,
    image_center=(4.5, 4.5),
    radius=0.5,
    n_points=20,
    write_obj=True,
    output_path="output/stent_comparison",
    interpolation_steps=28,
    bruteforce=False,
    sample_size=200,
)
```

This `from_array` function automatically aligns the frames within a pullback and then between pullbacks. The algorithm translates contours to the same centroid as the most proximal contour, and then finds the best rotation based on contour **AND** catheter points.

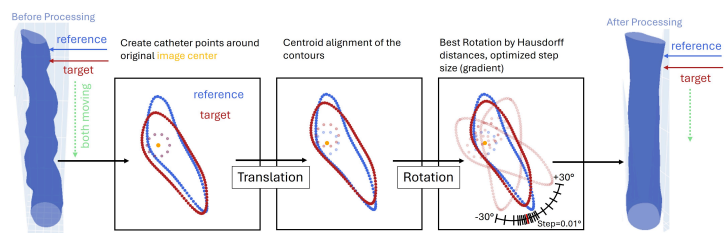

The number of catheter points ( `n_points` ) therefore influences how much weight is given to the original image center. For mostly round contours, where Hausdorff distances are similar in different angles, this image center can increase accuracy of the correct rotation. For stenotic sections or coronary artery anomalies, where the vessel has distinct shape difference, this number can be kept rather small (default 20 points compared to 500 for the contour).

`range_rotation_deg` and `step_rotation_deg` define the +/- degree range where the rotation is tested (default 90° so full range) and step_rotation_deg in what step sizes (default 0.5°). This algorithm is optimized and where it downsamples the original contour to 200 points, and performs coars steps (full provided range in 1° steps, then in +/- 5° degrees around the optimal angle in 0.1° steps and so on until the desired acccuracy). If bruteforce is set to 'True' the complete range is swept with the provided acccuracy (not recommended O(n^3)).

If `write_obj` is set to True, geometries will be saved as .obj files. if interpolation steps are not 0, additionally interpolated geometries will be created. This is useful if the dynamic behavior rendered later on. For example here a rendering of a non-aligned systolic stress-induced deformation in a coronary artery anomaly:

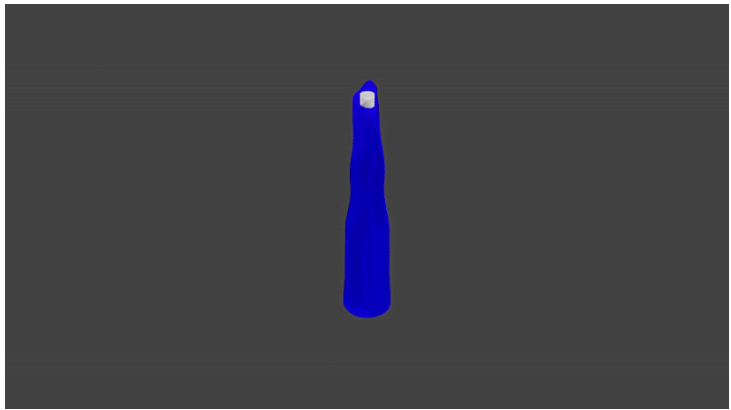

## 2. Alignment with a centerline

A centerline can be created directly from points. Points don't need any index, only x-, y- and z-coordinates:

| ... | ... | ... |
|-----|-----|-----|
| 12.6579 | -199.7824 | 1751.519 |
| 13.0847 | -200.3508 | 1751.8602 |
| 13.419 | -200.9894 | 1752.1491 |
| ... | ... | ... |

These could for example be stored in a .csv file and then be converted to a PyCenterline, which also includes the normals connecting the points:

```
cl_raw = np.genfromtxt("data/centerline_raw.csv", delimiter=",")
centerline = mm.numpy_to_centerline(cl_raw)
```

As soon as the centerline is created it will be automatically resampled to have the same spacing as the PyGeometry or PyGeometryPair, which will be aligned with the centerline.

This can either be done with three point alignment (preferred), where one point is corresponding to the reference point of the PyGeometry (e.g. aortic reference for coronary artery anomalies) and one point indicating the superior position and another point indicating the inferior position.

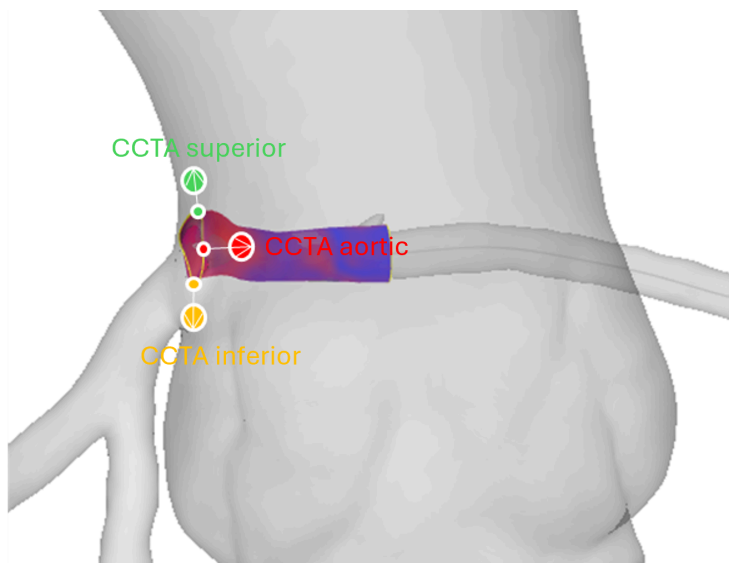

The reference contour is then best matched to these three points, all the leading points on the centerline are removed and the spacing is adjusted to match the z-spacing of the PyGeometry.

```
aligned_pair, cl_resampled = mm.to_centerline(
    mode="three_pt",
    centerline=centerline,
    geometry_pair=rest,                # e.g. Rest geometry (dia/sys)
    aortic_ref_pt=(12.26, -201.36, 1751.06),
    upper_ref_pt=(11.76, -202.19, 1754.80),
    lower_ref_pt=(15.66, -202.19, 1749.97)
)
```

## 3. Saving everything as .obj files

While every wrapper function allows to directly save the created geometries as .obj files (with optional interpolation), it is also possible to save any created geometry directly to an object file. The `to_obj` function can automatically detect the type of the object, and can be applied to PyGeometryPair, PyGeometry. For PyCenterline use the `centerline_to_obj` function.

```
mm.to_obj(aligned_pair.dia_geom, "data/aligned.obj")
mm.centerline_to_obj(cl_resampled, "data/resampled_cl.obj")
```

## 4. Utility functions to link to numpy

Any python object can be returned as numpy array, in case of PyGeometry and PyGeometryPair the different parts will be returned as a dictionary with their corresponding arrays (contours, catheters, walls, reference):

```
stress_dia_arr, stress_sys_arr = mm.to_array(stress)
aligned_arr = mm.to_array(aligned)
centerline_arr = mm.to_array(cl_resampled)
ostial_contour_arr = mm.to_array(rest.dia_geom.contours[-1])
```

Returns:

```
np.ndarray
    For PyContour or PyCenterline:
    A 2D array of shape (N, 4), where each row is (frame_index, x, y, z).

dict[str, np.ndarray]
    For PyGeometry:
    A dictionary with keys ["contours", "catheters", "walls", "reference"],
    each containing a 2D array of shape (M, 4), where M is the number of points in that layer.
    "reference" is a (1, 4) array or (0, 4) if missing.

Tuple[dict[str, np.ndarray], dict[str, np.ndarray]]
    For PyGeometryPair:
    A tuple of two dictionaries (one for diastolic, one for systolic), each in the same format
    as returned for a single PyGeometry.
```

## 5. Reordering algorithm

Especially in intravascular ultrasound imaging breathing can lead to additional bulk movements of frames due to relative catheter movement to the vessel. This can lead to complex patterns and the preferred solution is with the Option<record>. In this case algorithms can also be manually controlled. However, `multimodars` additionally provides a reordering algorithm that works by creating a cost matrix of Hausdorff distances between all frames in the geometry.

```
rest.reorder(delta=0.0, max_rounds=5)
```

## 6. Class methods

### PyContour

After creating a PyGeometry several utility methods provided. If a new contour is created from points and no centroid is available it can easily be calculated, additionally can the closest opposite points and the farthest points be identified:

```
contour.compute_centroid()
(p1, p2), distance = contour.find_closest_opposite()
(p1, p2), distance = contour.find_farthest_points()
```

For every contour the area and elliptic ratio can be returned. **CAVE:** units are calculated from the original image spacing, if contours were provided in pixels no meaningful result will be returned.

```
area = contour.get_area()
elliptic_ratio = contour.get_elliptic_ratio()
```

Contours can also be manipulated, however for additional safety operations are not performed in place but rather return a new contour that can then be set to the original position if needed.

```
contour = geometry.contours[2]
contour_rot = contour.rotate(20)
contour_trsl = contour_rot.translate((0.0, 1.0, 2.0))
geometry.set_cont(2, contour_trsl)
```

### PyGeometry/PyGeometryPair

The PyGeometry has some additional functionality, contours inside can be smoothed with a moving average and rotation and translation can be performed on Geometry level

```
geometry.smooth_contours(window_size=3)
geom_rot = geometry.rotate(20)
geom_trsl = geom_rot.translate((0.0, 1.0, 2.0))
```

Additionally there is a summary function to return minimal lumen area, maximum stenosis, and stenosis length in mm as a tuple for either PyGeometry or PyGeometryPair. For PyGeometryPair additionally a map with lumen area and elliptic ratio for either diastole and systole are provided. These results can then easily be translated to a numpy array.

```
geometries.get_summary()
geometries.dia_geom.get_summary()
geometries.sys_geom.get_summary()
# turn summary map to numpy array
_, deformation = geometries.get_summary()
deform_array = np.array(deformation)
```

Returns:

```
Geometry "Diastole":
MLA [mm²]: 5.57
Max. stenosis [%]: 58
Stenosis length [mm]: 2.99

Geometry "Systole":
MLA [mm²]: 4.71
Max. stenosis [%]: 69
Stenosis length [mm]: 11.20
```

| id | area_dia | ellip_dia | area_sys | ellip_sys | z |
|----|----------|-----------|----------|-----------|-------|
| 0 | 12.20 | 1.23 | 15.14 | 1.03 | 0.75 |
| 1 | 12.68 | 1.20 | 14.99 | 1.04 | 1.49 |
| 2 | 13.09 | 1.16 | 14.85 | 1.05 | 2.24 |
| 3 | 13.24 | 1.13 | 14.51 | 1.04 | 2.99 |
| 4 | 13.26 | 1.11 | 13.48 | 1.03 | 3.73 |
| 5 | 13.22 | 1.12 | 11.78 | 1.06 | 4.48 |
| 6 | 13.07 | 1.11 | 9.50 | 1.11 | 5.23 |
| 7 | 12.70 | 1.10 | 7.86 | 1.13 | 5.97 |
| 8 | 12.46 | 1.10 | 6.87 | 1.18 | 6.72 |
| 9 | 12.37 | 1.09 | 6.62 | 1.18 | 7.46 |
| 10 | 12.28 | 1.08 | 6.28 | 1.21 | 8.21 |
| 11 | 12.04 | 1.09 | 5.91 | 1.26 | 8.96 |
| 12 | 11.77 | 1.12 | 5.56 | 1.32 | 9.70 |
| 13 | 11.06 | 1.14 | 5.58 | 1.37 | 10.45 |
| 14 | 10.09 | 1.12 | 5.96 | 1.48 | 11.20 |
| 15 | 8.93 | 1.11 | 6.31 | 1.59 | 11.94 |
| 16 | 7.85 | 1.14 | 6.35 | 1.87 | 12.69 |
| 17 | 6.80 | 1.16 | 5.81 | 2.27 | 13.44 |
| 18 | 5.99 | 1.30 | 5.29 | 2.76 | 14.18 |
| 19 | 5.57 | 1.55 | 5.25 | 2.97 | 14.93 |
| 20 | 5.86 | 1.78 | 5.42 | 2.88 | 15.68 |
| 21 | 6.04 | 1.76 | 5.45 | 2.79 | 16.42 |
| 22 | 6.55 | 1.53 | 5.02 | 2.66 | 17.17 |
| 23 | 7.22 | 1.43 | 4.71 | 2.56 | 17.92 |

# API Reference

## API contents:

# Core Functions

---

**multimodars.align_manual**(*centerline, geometry_pair, rotation_angle, start_point, write=False, interpolation_steps=28, output_dir='output/aligned', case_name='None'*)

Creates centerline-aligned meshes for diastolic and systolic geometries based on three reference points (aorta, upper section, lower section). Only works for elliptic vessels e.g. coronary artery anomalies.

| | |
|---|---|
| **Parameters:** | • **centerline** – PyCenterline object |
| | • **geometry_pair** – PyGeometryPair object |
| | • **aortic_ref_pt** – Reference point for aortic position |
| | • **upper_ref_pt** – Upper reference point |
| | • **lower_ref_pt** – Lower reference point |
| | • **state** – Physiological state ("rest" or "stress") |
| | • **input_dir** – Input directory for raw geometries |
| | • **output_dir** – Output directory for aligned meshes |
| | • **interpolation_steps** – Number of interpolation steps |
| **Returns:** | PyGeometryPair |

**Example**

```
>>> import multimodars as mm
>>> dia, sys = mm.centerline_align(
...     "path/to/centerline.csv",
...     (1.0, 2.0, 3.0),
...     (4.0, 5.0, 6.0),
...     (7.0, 8.0, 9.0),
...     "rest"
... )
```

---

**multimodars.align_three_point**(*centerline, geometry_pair, aortic_ref_pt, upper_ref_pt, lower_ref_pt, angle_step_deg=1.0, write=False, interpolation_steps=28, output_dir='output/aligned', case_name='None'*)

Creates centerline-aligned meshes for diastolic and systolic geometries based on three reference points (aorta, upper section, lower section). Only works for elliptic vessels e.g. coronary artery anomalies.

| | |
|---|---|
| **Parameters:** | • **centerline** – PyCenterline object |
| | • **geometry_pair** – PyGeometryPair object |
| | • **aortic_ref_pt** – Reference point for aortic position |
| | • **upper_ref_pt** – Upper reference point |
| | • **lower_ref_pt** – Lower reference point |
| | • **state** – Physiological state ("rest" or "stress") |
| | • **input_dir** – Input directory for raw geometries |
| | • **output_dir** – Output directory for aligned meshes |
| | • **interpolation_steps** – Number of interpolation steps |
| **Returns:** | PyGeometryPair, PyCenterline (resampled) |

📖 ⑂ latest ▾

**Example**

```
>>> import multimodars as mm
>>> dia, sys = mm.align_three_point(
...     centerline,
...     geometry_pair,
...     (12.2605, -201.3643, 1751.0554),
...     (11.7567, -202.1920, 1754.7975),
...     (15.6605, -202.1920, 1749.9655),
... )
```

**multimodars.centerline_to_obj**(*cl, filename*)   [source]

**Write out a centerline as an OBJ with:**

- vertex positions (v x y z)
- vertex normals (vn nx ny nz), if normals are set
- a single poly-line (l 1 2 3 … N)

| Parameters: | • **cl** – A PyCenterline instance |
| | • **filename** (`str`) – Path to write (e.g. "my_centerline.obj") |
| Return type: | `None` |

**multimodars.create_catheter_geometry**(*geometry, image_center=Ellipsis, radius=0.5, n_points=20*)

Generate catheter contours and return a new PyGeometry with them filled in.

This function takes an existing PyGeometry, extracts all its contour points, computes catheter contours around those points, and returns a new PyGeometry with the catheter field populated.

| Parameters: | • **geometry** (-) |
| | • **image_center** (-) |
| | • **radius** (-) |
| | • **n_points** (-) |
| Return type: | A new `PyGeometry` with the same data but *catheter* field filled. |

**multimodars.from_array**(*mode, **kwargs*)   [source]

Unified entry for all array-based pipelines.

| Parameters: | • **mode** (*{"full", "doublepair", "singlepair", "single"}*) – Which array-based pipeline to run. |
| | • ****kwargs** (*dict*) – Keyword-only arguments required vary by mode (see below). |
| | • **Modes** (*Supported*) |
| | • --------------- |
| | • **"full"** (-) – |
| | from_array_full(rest_dia, rest_sys, stress_dia, stress_sys, |
| |     step_rotation_deg, range_rotation_deg, interpolation_steps, |
| |     rest_output_path, stress_output_path, diastole_output_path, |
| |     systole_output_path, image_center, radius, n_points) |
| | • **"doublepair"** (-) – |
| | from_array_doublepair(rest_dia, rest_sys, stress_dia, stress_sys, |

step_rotation_deg, range_rotation_deg, interpolation_steps,
rest_output_path, stress_output_path, image_center, radius, n_points)

- **"singlepair"** (-) –
  **from_array_singlepair(rest_dia, rest_sys, output_path,**

  step_rotation_deg, range_rotation_deg, interpolation_steps,
  image_center, radius, n_points)

- **"single"** (-) –
  **geometry_from_array(contours, walls, reference_point,**

  steps, range, image_center, radius, n_points, label, records, delta,
  max_rounds, diastole, sort, write_obj, output_path)

**Return type:** `Union` [ `Tuple` [ `PyGeometryPair` , `PyGeometryPair` , `PyGeometryPair` , `PyGeometryPair` , `list` , `list` , `list` , `list` ], `Tuple` [ `PyGeometryPair` , `PyGeometryPair` , `list` , `list` , `list` , `list` ], `Tuple` [ `PyGeometryPair` , `PyGeometryPair` , `list` , `list` ], `Tuple` [ `PyGeometry` , `list` ]]

**Returns:**

- Depends on *mode*
- - *"full"* –

  Tuple[PyGeometryPair, PyGeometryPair, PyGeometryPair,
  PyGeometryPair,

    list, list, list, list]

- - *"doublepair"* – Tuple[PyGeometryPair, PyGeometryPair, list, list, list, list]
- - *"singlepair"* – Tuple[PyGeometryPair, PyGeometryPair, list, list]
- - *"single"* – Tuple[PyGeometry, list]

**Raises:** **ValueError** – If an unsupported *mode* is passed.

---

multimodars.from_array_doublepair(*rest_geometry_dia*, *rest_geometry_sys*, *stress_geometry_dia*, *stress_geometry_sys*, *step_rotation_deg=0.5*, *range_rotation_deg=90.0*, *image_center=Ellipsis*, *radius=0.5*, *n_points=20*, *write_obj=True*, *rest_output_path='output/rest'*, *stress_output_path='output/stress'*, *interpolation_steps=28*, *bruteforce=False*, *sample_size=500*)

Processes two geometries in parallel.

Pipeline:

```
Rest:            Stress:
diastole          diastole
   |                 |
   v                 v
systole           systole
```

**Parameters:**

- **rest_geometry_dia** (-)
- **rest_geometry_sys** (-)
- **stress_geometry_dia** (-)
- **stress_geometry_sys** (-)
- **degree.** (- *step_rotation_deg (default 0.5°)* – Rotation step in)
- **degree.** (- *range_rotation_deg (default 90°)* – Rotation (+/-) range
- **180°.** (*for 90° total range*)
- **true)** (- *write_obj (default)*

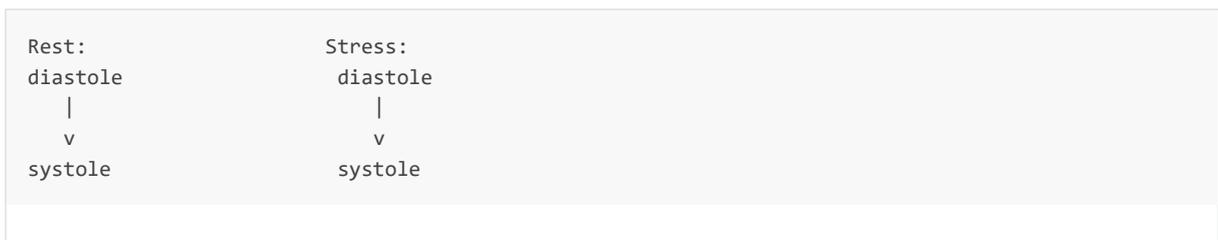

**Returns:**

- **(default** (- *interpolation_steps*)
- **(default**
- **(default**
- **false)** (- *bruteforce (default)*
- **to** (- *sample_size (default 200) number of points to downsample*)
-
- A tuple `(rest_pair, stress_pair)` of type `(PyGeometryPair, PyGeometryPair)` ,
- *containing the interpolated diastole/systole geometries for REST and STRESS.*
- A 4-tuple of `Vec<id, matched_to, rel_rot_deg, total_rot_deg, tx, ty, centroid_x, centroid_y>`
- *for (diastole logs, systole logs, diastole stress logs, systole stress logs).*

**Raises:**    **RuntimeError` if any Rust-side processing fails** –

## Example

```python
import multimodars as mm
rest_pair, stress_pair, _ = mm.from_array_doublepair(
    rest_dia, rest_sys,
    stress_dia, stress_sys,
    steps_best_rotation=0.2,
    interpolation_steps=32,
    rest_output_path="out/rest",
    stress_output_path="out/stress"
)
```

---

**multimodars.from_array_full**(*rest_geometry_dia, rest_geometry_sys, stress_geometry_dia, stress_geometry_sys, step_rotation_deg=0.5, range_rotation_deg=90.0, image_center=Ellipsis, radius=0.5, n_points=20, write_obj=True, rest_output_path='output/rest', stress_output_path='output/stress', diastole_output_path='output/diastole', systole_output_path='output/systole', interpolation_steps=28, bruteforce=False, sample_size=200*)

Process four existing `PyGeometry` objects (rest-dia, rest-sys, stress-dia, stress-sys) in parallel, aligning and interpolating between phases.

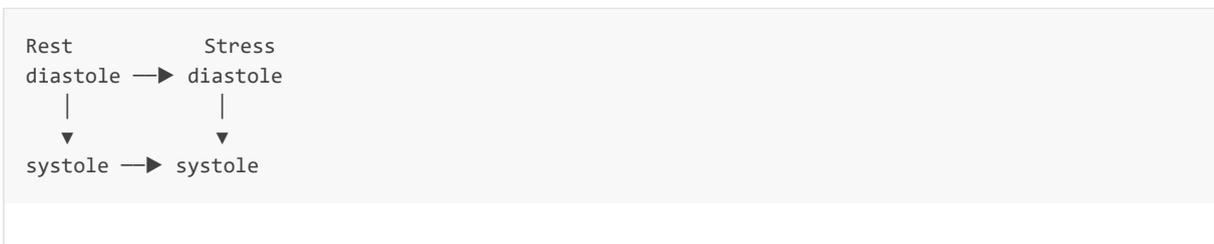

**Parameters:**

- **rest_geometry_dia** (-)
- **rest_geometry_sys** (-)
- **stress_geometry_dia** (-)
- **stress_geometry_sys** (-)
- **degree** (- *range_rotation_deg (default 90°) – Rotation (+/-) range in*)
- **degree**
- **180°** (*for 90° total range*)
- **true)** (- *write_obj (default)*)
- **(default** (- *interpolation_steps*)
- **(default**
- **(default**

- **(default
- **(default
- **false)** (- *bruteforce (default)*
- **to** (- *sample_size (default 200) number of points to downsample)*

**Returns:**

- 
- A `PyGeometryPair` for rest, stress, diastole, systole.
- A 4-tuple of `Vec<id, matched_to, rel_rot_deg, total_rot_deg, tx, ty, centroid_x, centroid_y>`
- *for (diastole logs, systole logs, diastole stress logs, systole stress logs).*

**Raises:** **RuntimeError`** if any Rust-side processing fails –

### Example

```python
import multimodars as mm
# Assume you have four PyGeometry objects from earlier:
rest, stress, dia, sys, _ = mm.from_array_full(
    rest_dia, rest_sys, stress_dia, stress_sys,
    steps_best_rotation=0.1,
    interpolation_steps=28,
    rest_output_path="out/rest",
    stress_output_path="out/stress"
)
rest_pair, stress_pair, dia_pair, sys_pair = full
```

**multimodars.from_array_singlepair**(*geometry_dia, geometry_sys, step_rotation_deg=0.5, range_rotation_deg=90.0, image_center=Ellipsis, radius=0.5, n_points=20, write_obj=True, output_path='output/singlepair', interpolation_steps=28, bruteforce=False, sample_size=500*)

Interpolate between two existing `PyGeometry` objects (diastole and systole) and return a `PyGeometryPair` containing both phases.

```
Geometry pipeline:
  diastole ──▶ systole
```

**Parameters:**

- **geometry_dia** (-)
- **geometry_sys** (-)
- **degree** (- *range_rotation_deg (default 90°) – Rotation (+/-) range in)*
- **degree**
- **180°** (*for 90° total range)*
- **true)** (- *write_obj (default)*
- **output_path** (-)
- **(default** (- *interpolation_steps)*
- **false)** (- *bruteforce (default)*
- **to** (- *sample_size (default 200) number of points to downsample)*

**Returns:**

- 
- A `PyGeometryPair` tuple containing the diastole and systole geometries with interpolation applied.
- A 2-tuple of `Vec<id, matched_to, rel_rot_deg, total_rot_deg, tx, ty, centroid_x, centroid_y>`

- *for (diastole logs, systole logs).*

**Raises:**     **RuntimeError` if the underlying Rust function fails** –

## Example

```python
import multimodars as mm
pair, _ = mm.from_array_singlepair(
    rest_dia, rest_sys,
    output_path="out/single",
    steps_best_rotation=0.1,
    interpolation_steps=30
)
```

**multimodars.from_file**(*mode, \*\*kwargs*)     [source]

A unified entrypoint for all *from_file_** variants.

**Parameters:**     • **mode** (*{"full","doublepair","singlepair","single"}*) –
Selects which low-level function is called.

  ○ **"full" → needs rest_input_path, stress_input_path,**

     rest_output_path, stress_output_path, diastole_output_path,
     systole_output_path, step_rotation_deg, range_rotation_deg,
     interpolation_steps, image_center, radius, n_points

  ○ **"doublepair" → needs rest_input_path, stress_input_path,**

     rest_output_path, stress_output_path, step_rotation_deg,
     range_rotation_deg, interpolation_steps, image_center, radius,
     n_points

  ○ **"singlepair" → needs input_path, output_path,**

     step_rotation_deg, range_rotation_deg, interpolation_steps,
     image_center, radius, n_points

  ○ **"single" → needs input_path, output_path,**

     step_rotation_deg, range_rotation_deg, diastole, image_center,
     radius, n_points

• **\*\*kwargs** (*dict*) – Keyword arguments required depend on *mode* (see
above).

**Return type:**     `Union` [ `Tuple` [ `PyGeometryPair` , `PyGeometryPair` , `PyGeometryPair` , `PyGeometryPair` ,
`list` , `list` , `list` , `list` ], `Tuple` [ `PyGeometryPair` , `PyGeometryPair` , `list` ,
`list` , `list` , `list` ], `Tuple` [ `PyGeometryPair` , `PyGeometryPair` , `list` , `list` ],
`Tuple` [ `PyGeometryPair` , `list` ]]

**Returns:**

• *Union[* – # full now returns 4 geometry pairs *and* 4 log-lists Tuple[
PyGeometryPair, PyGeometryPair, PyGeometryPair, PyGeometryPair, list,
list, list, list, ], # doublepair returns 2 geom + 4 log-lists Tuple[
PyGeometryPair, PyGeometryPair, list, list, list, list, ], # singlepair returns 1
geom-pair + 2 log-lists Tuple[ PyGeometryPair, PyGeometry
# single returns 1 geom + 1 log-list Tuple[ PyGeometryPair, list, ]

• *]* – The exact return shape depends on *mode.*

**Raises:**    **ValueError** – If an unsupported *mode* is passed.

---

**multimodars.from_file_doublepair**(*rest_input_path, stress_input_path, step_rotation_deg=0.5, range_rotation_deg=90.0, image_center=Ellipsis, radius=0.5, n_points=20, write_obj=True, rest_output_path='output/rest', stress_output_path='output/stress', interpolation_steps=28, bruteforce=False, sample_size=500*)

Processes two geometries in parallel.

Pipeline:

```
Rest:                Stress:
diastole              diastole
   |                     |
   v                     v
systole               systole
```

Arguments:

- `rest_input_path` – Path to REST input file
- `stress_input_path` – Path to STRESS input file
- `step_rotation_deg` (default 0.5°) – Rotation step in degree
- `range_rotation_deg` (default 90°) – Rotation (+/-) range in degree, for 90° total range 180°
- `image_center` (default (4.5mm, 4.5mm)) in mm
- `radius` (default 0.5mm) in mm for catheter
- `n_points` (default 20) number of points for catheter, more points stronger influence of image center
- `write_obj` (default true)
- `rest_output_path` (default "output/rest")
- `stress_output_path` (default "output/stress")
- `interpolation_steps` (default 28)
- `bruteforce` (default false)
- `sample_size` (default 200) number of points to downsample to

CSV format:

```
Frame Index, X-coord (mm), Y-coord (mm), Z-coord (mm)
185, 5.32, 2.37, 0.0
...
```

Returns:

A `PyGeometryPair` for rest, stress. A 4-tuple of `Vec<id, matched_to, rel_rot_deg, total_rot_deg, tx, ty, centroid_x, centroid_y>` for (diastole logs, systole logs, diastole stress logs, systole stress logs).

Example:

```python
import multimodars as mm
rest, stress, _ = mm.from_file_doublepair(
    "data/ivus_rest", "data/ivus_stress"
)
```

**multimodars.from_file_full**(*rest_input_path, stress_input_path, step_rotation_deg=0.5, range_rotation_deg=90.0, image_center=Ellipsis, radius=0.5, n_points=20, write_obj=True, rest_output_path='output/rest', stress_output_path='output/stress', diastole_output_path='output/diastole', systole_output_path='output/systole', interpolation_steps=28, bruteforce=False, sample_size=500*)

Processes four geometries in parallel.

Pipeline:

```
Rest:                   Stress:
diastole  ------------> diastole
   |                       |
   v                       v
systole   ------------> systole
```

Arguments:

- `rest_input_path` – Path to REST input file
- `stress_input_path` – Path to STRESS input file
- `step_rotation_deg` (default 0.5°) – Rotation step in degree
- `range_rotation_deg` (default 90°) – Rotation (+/-) range in degree, for 90° total range 180°
- `image_center` (default (4.5mm, 4.5mm)) in mm
- `radius` (default 0.5mm) in mm for catheter
- `n_points` (default 20) number of points for catheter, more points stronger influence of image center
- `write_obj` (default true)
- `rest_output_path` (default "output/rest")
- `stress_output_path` (default "output/stress")
- `diastole_output_path` (default "output/diastole")
- `systole_output_path` (default "output/systole")
- `interpolation_steps` (default 28)
- `bruteforce` (default false)
- `sample_size` (default 200) number of points to downsample to

CSV format:

```
Frame Index, X-coord (mm), Y-coord (mm), Z-coord (mm)
185, 5.32, 2.37, 0.0
...
```

Returns:

A `PyGeometryPair` for rest, stress, diastole, systole. A 4-tuple of `Vec<id, matched_to, rel_rot_deg, total_rot_deg, tx, ty, centroid_x, centroid_y>` for (diastole logs, systole logs, diastole stress logs, systole stress logs).

Example:

```python
import multimodars as mm
rest, stress, dia, sys, _ = mm.from_file_full(
    "data/ivus_rest", "data/ivus_stress"
)
```

**multimodars.from_file_single**(*input_path, step_rotation_deg=0.5, range_rotation_deg=90.0, diastole=True, image_center=Ellipsis, radius=0.5, n_points=20, write_obj=True, output_path='output/single', bruteforce=False, sample_size=500*)

Processes a single geometry (either diastole or systole) from an IVUS CSV file.

```
Rest/Stress pipeline (choose phase via `diastole` flag):
  e.g. diastole
```

**Parameters:**
- **input_path** (-)
- **degree** (- *range_rotation_deg (default 90°) – Rotation (+/-) range in*)
- **degree**
- **180°** (*for 90° total range*)
- **true)** (- *write_obj (default*)
- **(4.5mm** (- *image_center (default*)
- **4.5mm))** (*(x, y) center for processing.*)
- **(default** (- *output_path*)
- **(default**
- **true)**
- **(default**
- **false)** (- *bruteforce (default*)
- **to** (- *sample_size (default 200) number of points to downsample*)

**Returns:**
-
- A `PyGeometry` containing the processed contour for the chosen phase.
- A `Vec<id, matched_to, rel_rot_deg, total_rot_deg, tx, ty, centroid_x, centroid_y>`.

**Raises:** **RuntimeError` if the underlying Rust pipeline fails** –

**Example**

```python
import multimodars as mm
geom, _ = mm.from_file_single(
    "data/ivus.csv",
    steps_best_rotation=0.5,
    range_rotation_rad=90,
    output_path="out/single",
    diastole=False
)
```

**multimodars.from_file_singlepair**(*input_path, step_rotation_deg=0.5, range_rotation_deg=90.0, image_center=Ellipsis, radius=0.5, n_points=20, write_obj=True, output_path='output/singlepair', interpolation_steps=28, bruteforce=False, sample_size=500*)

Processes two geometries (rest and stress) in parallel from an input CSV, returning a single `PyGeometryPair` for the chosen phase.

```
Rest/Stress pipeline:
    diastole
       |
       ▼
    systole
```

**Parameters:**
- **input_path** (-)
- **degree** (- *range_rotation_deg (default 90°) – Rotation (+/-) range in*)
- **degree**
- **180°** (*for 90° total range*)
- **(4.5mm** (- *image_center (default*)
- **4.5mm))** (*Center coordinates (x, y).*)
- **radius.** (- *radius (default 0.5mm) Processing*)
- **points.** (- *n_points (default 20) Number of boundary*)
- **true)** (- *write_obj (default*)
- **output_path** (-)
- **steps.** (- *interpolation_steps (default 28) Number of interpolation*)
- **false)** (- *bruteforce (default*)
- **to** (- *sample_size (default 200) number of points to downsample*)
- **Format** (*CSV*)
- **----------**
- **is** (*The CSV must have no header. Each row*)
- **code-block:** (..) – text: 185, 5.32, 2.37, 0.0 ...

**Returns:**
- 
- A single `PyGeometryPair` for (rest or stress) geometry.
- A 2-tuple of `Vec<id, matched_to, rel_rot_deg, total_rot_deg, tx, ty, centroid_x, centroid_y>`
- *for (diastole logs, systole logs).*

**Raises:**   **RuntimeError` if the Rust pipeline fails** –

**Example**

```python
import multimodars as mm
pair, _ = mm.from_file_singlepair(
    "data/ivus_rest.csv",
    "output/rest"
)
```

This is a thin Python wrapper around the Rust implementation.

**multimodars.geometry_from_array**(*geometry, step_rotation_deg=0.5, range_rotation_deg=90.0, image_center=Ellipsis, radius=0.5, n_points=20, label='None', records=None, delta=0.0, max_rounds=5, diastole=True, sort=True, write_obj=False, output_path='output/single', bruteforce=False, sample_size=500*)

Process an existing `PyGeometry` by optionally reordering, aligning, and refining its contours, walls, and catheter data based on various criteria.

This wraps the internal Rust function `geometry_from_array_rs`, which: 1. Builds catheter contours (if `n_points > 0`), 2. Optionally reorders contours using provided `records` and z-coordinate sorting, 3. Aligns frames and refines the contour ordering via dynamic programming or 2-opt, 4. Smooths the final geometry, 5. Optionally writes OBJ meshes.

**Parameters:**
- **geometry** (-)
- **degree** (- *range_rotation_deg (default 90°) – Rotation (+/-) range in*)

- **degree**
- **180°** (*for 90° total range*)
- (**default** (- *output_path*)
- (**default**
- (**default**
- (**default**
- (**default**
- (**default**
- (**default**
- (**default**
- (**default**
- (**default**
- (**default**
- **false**) (- *bruteforce (default)*
- **to** (- *sample_size (default 200) number of points to downsample*)

**Returns:**

- 
- A new `PyGeometry` instance containing reordered, aligned, and smoothed contours.
- A `Vec<id, matched_to, rel_rot_deg, total_rot_deg, tx, ty, centroid_x, centroid_y>`.

**Raises:** RuntimeError` if any Rust-side processing step fails –

**Example**

```python
import multimodars as mm
# Suppose ``geo`` is an existing PyGeometry from earlier processing:
refined, _ = mm.geometry_from_array(
    geo,
    steps_best_rotation=200,
    range_rotation_rad=1.0,
    records=my_records,
    delta=0.2,
    max_rounds=3,
    sort=True,
    write_obj=True,
    output_path="out/mesh"
)
```

---

**multimodars.numpy_to_centerline**(*arr, aortic=False*)    [source]

Build a PyCenterline from a numpy array of shape (N,3), where each row is (x, y, z).

**Parameters:**
- **arr** ( `ndarray` ) – np.ndarray of shape (N,3)
- **aortic** ( `bool` ) – whether to mark each point as aortic

**Return type:** `PyCenterline`

**Returns:** PyCenterline

---

**multimodars.numpy_to_geometry**(*contours_arr, catheters_arr, walls_arr, reference_arr*)    [source]

Build a PyGeometry from four (M, 4) NumPy arrays or structured arrays, one per l
by frame_index.

Each row in the `*_arr` is [frame_index, x, y, z].

| Return type: | `PyGeometry` |
| Parameters: | |

- **contours_arr** (*ndarray*)
- **catheters_arr** (*ndarray*)
- **walls_arr** (*ndarray*)
- **reference_arr** (*ndarray*)

**Returns a PyGeometry containing:**

- contours: list of PyContour (one per frame in contours_arr)
- catheters: list of PyContour (one per frame in catheters_arr)
- walls: list of PyContour (one per frame in walls_arr)
- reference: single PyContourPoint from reference_arr[0]

---

**multimodars.to_array**(*generic*)        [source]

Convert various multimodars Py* objects into numpy array(s) or dictionaries of arrays.

| Parameters: | **generic** (*PyContour*, *PyCenterline*, *PyGeometry*, *or PyGeometryPair*) – The object to be converted to numpy representation. |
| Return type: | `Union [ ndarray , dict , Tuple [ dict , dict ]]` |
| Returns: | |

- *np.ndarray* – For PyContour or PyCenterline: A 2D array of shape (N, 4), where each row is (frame_index, x, y, z).
- *dict[str, np.ndarray]* – For PyGeometry: A dictionary with keys ["contours", "catheters", "walls", "reference"], each containing a 2D array of shape (M, 4), where M is the number of points in that layer. "reference" is a (1, 4) array or (0, 4) if missing.
- *Tuple[dict[str, np.ndarray], dict[str, np.ndarray]]* – For PyGeometryPair: A tuple of two dictionaries (one for diastolic, one for systolic), each in the same format as returned for a single PyGeometry.

| Raises: | **TypeError** – If the input type is not one of the supported multimodars types. |

---

**multimodars.to_centerline**(*mode, \*\*kwargs*)        [source]

Unified entry for all to_centerline pipelines.

## Supported modes

::

- **"three_pt" → align_three_point(centerline, geometry_pair, aortic_ref_pt, upper_ref_pt,**
  lower_ref_pt, angle_step, write, interpolation_steps, output_dir, case_name)

- **"manual" → align_manual(centerline, geometry_pair, rotation_angle, start_point,**
  write, interpolation_steps, output_dir, case_name)

| type mode: | `Literal [ 'three_pt' , 'manual' ]` |
| param mode: | Which array-based pipeline to run (see "Supported modes" above). |
| type mode: | {"three_pt","manual"} |
| type \*\*kwargs: | `Any` |
| param \*\*kwargs: | Keyword-only arguments required vary by mode (see above). |
| type \*\*kwargs: | dict |

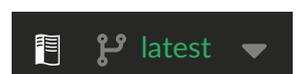

| | |
|---|---|
| **returns:** | Depends on *mode*. |
| **rtype:** | PyGeometryPair |
| **raises ValueError:** | If an unsupported *mode* is passed. |
| **Parameters:** | • **mode** (*Literal['three_pt', 'manual']*) |
| | • **kwargs** (*Any*) |
| **Return type:** | *Tuple*[PyGeometryPair, PyCenterline] |

**multimodars.to_obj**(*geometry, output_path, contours=True, walls=True, catheter=True, filename_contours='contours.obj', material_contours='contours.mtl', filename_catheter='catheter.obj', material_catheter='catheter.mtl', filename_walls='walls.obj', material_walls='walls.mtl'*)

Convert a `PyGeometry` object into one or more OBJ files and write them to disk.

This function takes a Python-exposed geometry ( `PyGeometry` ), converts it into the corresponding Rust geometry, and writes the specified components (contours, walls, catheter) as OBJ meshes without UV coordinates. Each component is written to its own file, with a corresponding MTL material file.

# Arguments

- `geometry` - Input `PyGeometry` instance containing the mesh data.
- `output_path` - Directory path where the OBJ and MTL files will be written.
- `contours` - Whether to export the contour mesh (default: `true` ).
- `walls` - Whether to export the wall mesh (default: `true` ).
- `catheter` - Whether to export the catheter mesh (default: `true` ).
- `filename_contours` - Filename for the contour OBJ (default: "contours.obj").
- `material_contours` - Filename for the contour MTL (default: "contours.mtl").
- `filename_catheter` - Filename for the catheter OBJ (default: "catheter.obj").
- `material_catheter` - Filename for the catheter MTL (default: "catheter.mtl").
- `filename_walls` - Filename for the walls OBJ (default: "walls.obj").
- `material_walls` - Filename for the walls MTL (default: "walls.mtl").

# Errors

Returns a *PyRuntimeError* if any of the underlying file writes fail.

# Classes

## API contents:

- PyContourPoint
  - `PyContourPoint`

- PyContour
  - `PyContour`

- PyGeometry
  - `PyGeometry`

- PyGeometryPair
  - `PyGeometryPair`

- PyRecord
  - `PyRecord`

- PyCenterlinePoint
  - `PyCenterlinePoint`

- PyCenterline
  - `PyCenterline`

# PyContourPoint

*class* **multimodars.PyContourPoint**(*frame_index, point_index, x, y, z, aortic*)

Bases: `object`

Python representation of a 3D contour point

### `frame_index`

Frame number in sequence

**Type:** int

### `point_index`

Index within contour

**Type:** int

### `x`

X-coordinate in mm

**Type:** float

### `y`

Y-coordinate in mm

**Type:** float

### `z`

Z-coordinate (depth) in mm

**Type:** float

### `aortic`

Flag indicating aortic position (in case of intramural course)

**Type:** bool

**Example**

```
>>> point = PyContourPoint(
...     frame_index=0,
...     point_index=1,
...     x=1.23,
...     y=4.56,
...     z=7.89,
...     aortic=True
... )
```

**distance**(*other*)

Euclidean distance to another PyContourPoint

**Parameters:** **point** (*PyContourPoint*) – Any other PyContourPoint.

**Example**

```
>>> p1.distance(p2)
```

# PyContour

*class* `multimodars.PyContour(id, points)`

Bases: `object`

Python representation of a 3D contour

`id`

Contour number in sequence

**Type:** int

`points`

Vector of ContourPoints

**Type:** [PyContourPoint]

`centroid`

Tuple containing x-, y-, z-coordinates

**Type:** float, float, float

**Example**

```
>>> contour = PyContour(
...     id=0,
...     points=[point1, point2, ...],
...     centroid=(1.0, 1.0, 1.0)
... )
```

`compute_centroid()`

Calculates the contours centroid by averaging over all coordinates

**Example**

```
>>> contour.compute_centroid()
```

`find_closest_opposite()`

Finds closest points on opposite sides of the contour

**Returns:** Pair of points and their Euclidean distance
**Return type:** Tuple[Tuple[PyContourPoint, PyContourPoint], float]

**Example**

```
>>> (p1, p2), distance = contour.find_closest_opposite()
```

**find_farthest_points**()

Finds the two farthest points in the contour

| Returns: | Pair of points and their Euclidean distance |
|---|---|
| Return type: | Tuple[Tuple[PyContourPoint, PyContourPoint], float] |

**Example**

```
>>> (p1, p2), distance = contour.find_farthest_points()
```

**get_area**()

Get the area of the current contour using shoelace formula

| Returns: | Area of the current contour in the unit that the original contour data was provided (e.g. mm2). |
|---|---|
| Return type: | float |

**Example**

```
>>> area = contour.get_area()
```

**get_elliptic_ratio**()

Get the elliptic ratio of the current contour

| Returns: | Ratio of farthest points distance divided by closest opposite points distance. |
|---|---|
| Return type: | float |

**Example**

```
>>> elliptic_ratio = contour.get_elliptic_ratio()
```

**points_as_tuples**()

Returns contour points as list of (x, y, z) tuples

**Example**

```
>>> contour.points_as_tuples()
[(1.0, 2.0, 3.0), (4.0, 5.0, 6.0)]
```

**rotate**(*angle_deg*)

Rotate a given contour around it's own centroid by an angle in degrees.

| Returns: | Original Contour rotated around it's centroid |
|---|---|
| Return type: | PyContour |

**Example**

```
>>> contour = contour.rotate(20)
```

**sort_contour_points**()

Sort points within a contour, so highest y-coord point has index 0 and all the others are sorted counterclockwise

| | |
|---|---|
| **Returns:** | Original Contour rearranged points.point_idx |
| **Return type:** | PyContour |

**Example**

```
>>> contour = contour.sort_contour_points()
```

**translate**(*dx*, *dy*, *dz*)

translate a given contour by x, y, z coordinates

| | |
|---|---|
| **Parameters:** | • **dx** (*float*) – Translation in x-direction. |
| | • **dy** (*float*) – Translation in y-direction. |
| | • **dz** (*float*) – Translation in z-direction. |
| **Returns:** | Original Contour translated to (x, y, z) |
| **Return type:** | PyContour |

**Example**

```
>>> contour = contour.translate((0.0, 1.0, 2.0))
```

# PyGeometry

*class* `multimodars.PyGeometry(`*contours, catheters, walls, reference_point*`)`

Bases: `object`

Python representation of a full geometry set

**Contains:**

- Vessel contours
- Catheter points
- Wall contours
- Reference point

`contours`

Vessel contours

> **Type:** List[PyContour]

`catheter`

Catheter points

> **Type:** List[PyContour]

`walls`

Wall contours

> **Type:** List[PyContour]

`reference_point`

Reference position

> **Type:** PyContourPoint

**Example**

```
>>> geom = PyGeometry(
...     contours=[contour1, contour2],
...     catheter=[catheter_points],
...     walls=[wall1, wall2],
...     reference_point=ref_point
... )
```

`get_summary()`

Get a compact summary of lumen properties for this geometry.

> **Returns:**
>
> (mla, max_stenosis, stenosis_length_mm)

latest ▾

mla (float): minimal lumen area (same units as contour.area(), e.g. mm^2) max_stenosis (float): 1 - (mla / biggest_area)
stenosis_length_mm (float): length (in mm) of the longest contiguous region where contour area < threshold.

**Return type:** tuple

**Threshold logic (implemented by assumption):**

If ALL contours have elliptic_ratio < 1.3 we treat the vessel as "elliptic" and use a more lenient threshold of 0.70 * biggest_area. Otherwise we use a stricter threshold of 0.50 * biggest_area (50%).

**reorder**(*delta*, *max_rounds*)

Re-orders and realigns the sequence of contours to minimize a combined spatial + index-jump cost.

**Parameters:**
- **delta** (*float*) – Jump penalty weight between contour IDs.
- **max_rounds** (*int*) – Maximum refinement iterations.
- **steps** (*int*) – Number of steps for frame alignment.
- **range** (*float*) – Range parameter for frame alignment.

**Returns:** A new geometry with contours and catheter re-ordered and aligned.

**Return type:** PyGeometry

**rotate**(*angle_deg*)

Rotate all contours/walls/catheters of a given geometry around it's own centroid by an angle in degrees. Catheters are rotated around the same centroid as contour.

**Returns:** Original Geometry rotated around it's centroid

**Return type:** PyGeometry

**Example**

```
>>> geometry = geometry.rotate(20)
```

**set_catheter**(*idx*, *catheter*)

Replace the contour at *idx* (can be negative).

**Parameters:**
- **idx** (*float*) – Target index to replace.
- **catheter** (*PyContour*) – Catheter to set to target index.

**Example**

```
>>> catheter = geom.catheters[0].copy()
>>> geom.set_catheter(10, catheter)
```

**set_contour**(*idx*, *contour*)

Replace the contour at *idx* (can be negative).

**Parameters:**
- **idx** (*float*) – Target index to replace.

- **contour** (*PyContour*) – Contour to set to target index.

**Example**

```
>>> contour = geom.contours[0].copy()
>>> geom.set_contour(10, contour)
```

**set_wall**(*idx*, *wall*)

Replace the contour at *idx* (can be negative).

Parameters: 
- **idx** (*float*) – Target index to replace.
- **wall** (*PyContour*) – Wall-contour to set to target index.

**Example**

```
>>> wall = geom.walls[0].copy()
>>> geom.set_wall(10, wall)
```

**smooth_contours**()

Applies smoothing to all contours using a threepoint moving average

**Example**

```
>>> geom.smooth_contours()
```

**translate**(*dx*, *dy*, *dz*)

Translates all contours, walls, and catheters in a geometry by (dx, dy, dz).

Parameters: 
- **dx** (*float*) – translation in x-direction.
- **dy** (*float*) – translation in y-direction.
- **dz** (*float*) – translation in z-direction.

Returns: A new PyGeometry with all elements translated.

# PyGeometryPair

*class* `multimodars.PyGeometryPair(dia_geom, sys_geom)`

Bases: `object`

Python representation of a diastolic/systolic geometry pair

`dia_geom`

> Diastolic geometry
>
> > **Type:** PyGeometry

`sys_geom`

> Systolic geometry
>
> > **Type:** PyGeometry

**Example**

```
>>> pair = PyGeometryPair(
...     dia_geom=diastole,
...     sys_geom=systole
... )
```

`create_deformation_table()`

`get_summary()`

> Get summaries for both diastolic and systolic geometries.
>
> Returns a tuple of: ((dia_mla, dia_max_stenosis, dia_len_mm), (sys_mla, sys_max_stenosis, sys_len_mm)) and a matrix (N, 6): (contour id, area_dia, ellip_dia, area_sys, ellip_sys, z-coordinate)
>
> This calls `get_summary()` on each contained PyGeometry and returns both results. and additionally assesses dynamic between the two PyGeometry object (area, elliptic ratio)

latest

# PyRecord

*class* **multimodars.PyRecord**(*frame, phase, measurement_1, measurement_2*)

Bases: `object`

Python representation of a measurement record

**frame**

Frame number

**Type:** int

**phase**

Cardiac phase ('D'/'S') for diastole or systole

**Type:** str

**measurement_1**

Primary measurement. In coronary artery anomalies thickness between aorta and coronary.

**Type:** float, optional

**measurement_2**

Secondary measurement. In coronary artery anomalies thickness between pulmonary artery and coronary.

**Type:** float, optional

**Example**

```
>>> record = PyRecord(
...     frame=5,
...     phase="D",
...     measurement_1=1.4,
...     measurement_2=2.1
... )
```

# PyCenterlinePoint

*class* `multimodars.PyCenterlinePoint`(*contour_point, normal*)

Bases: `object`

Python representation of a centerline point

Combines a contour point with its normal vector

`contour_point`

Position in 3D space

**Type:** [PyContourPoint](#)

`normal`

Normal vector (nx, ny, nz)

**Type:** Tuple[float, float, float]

**Example**

```
>>> cl_point = PyCenterlinePoint(
...     contour_point=point,
...     normal=(0.0, 1.0, 0.0)
... )
```

# PyCenterline

*class* `multimodars.PyCenterline`(*points*)

Bases: `object`

Python representation of a vessel centerline

`points`

    Ordered points along centerline

        **Type:**    List[PyCenterlinePoint]

**Example**

```
>>> centerline = PyCenterline(points=[p1, p2, p3])
```

*static* `from_contour_points`(*contour_points*)

    Build a Centerline from a flat list of PyContourPoint.

        **Parameters:**    contour_points (*List[PyContourPoint]*) – sequence of points in order.
        **Returns:**    PyCenterline

**Example**

```
>>> pts = [PyContourPoint(...), PyContourPoint(...), ...]
>>> cl = PyCenterline.from_contour_points(pts)
```

`points_as_tuples`()

# Processing Functions

**multimodars.from_file_full**(*rest_input_path, stress_input_path, step_rotation_deg=0.5, range_rotation_deg=90.0, image_center=Ellipsis, radius=0.5, n_points=20, write_obj=True, rest_output_path='output/rest', stress_output_path='output/stress', diastole_output_path='output/diastole', systole_output_path='output/systole', interpolation_steps=28, bruteforce=False, sample_size=500*)

Processes four geometries in parallel.

Pipeline:

```
Rest:                     Stress:
diastole  ------------->  diastole
    |                         |
    v                         v
systole   ------------->  systole
```

Arguments:

- `rest_input_path` – Path to REST input file
- `stress_input_path` – Path to STRESS input file
- `step_rotation_deg` (default 0.5°) – Rotation step in degree
- `range_rotation_deg` (default 90°) – Rotation (+/-) range in degree, for 90° total range 180°
- `image_center` (default (4.5mm, 4.5mm)) in mm
- `radius` (default 0.5mm) in mm for catheter
- `n_points` (default 20) number of points for catheter, more points stronger influence of image center
- `write_obj` (default true)
- `rest_output_path` (default "output/rest")
- `stress_output_path` (default "output/stress")
- `diastole_output_path` (default "output/diastole")
- `systole_output_path` (default "output/systole")
- `interpolation_steps` (default 28)
- `bruteforce` (default false)
- `sample_size` (default 200) number of points to downsample to

CSV format:

```
Frame Index, X-coord (mm), Y-coord (mm), Z-coord (mm)
185, 5.32, 2.37, 0.0
...
```

Returns:

A `PyGeometryPair` for rest, stress, diastole, systole. A 4-tuple of `Vec<id, matched_to, total_rot_deg, tx, ty, centroid_x, centroid_y>` for (diastole logs, systole logs, diastole stress logs, systole stress logs).

Example:

```python
import multimodars as mm
rest, stress, dia, sys, _ = mm.from_file_full(
    "data/ivus_rest", "data/ivus_stress"
)
```

**multimodars.from_file_doublepair**(*rest_input_path, stress_input_path, step_rotation_deg=0.5, range_rotation_deg=90.0, image_center=Ellipsis, radius=0.5, n_points=20, write_obj=True, rest_output_path='output/rest', stress_output_path='output/stress', interpolation_steps=28, bruteforce=False, sample_size=500)*

Processes two geometries in parallel.

Pipeline:

```
Rest:              Stress:
diastole            diastole
  |                   |
  v                   v
systole             systole
```

Arguments:

- `rest_input_path` – Path to REST input file
- `stress_input_path` – Path to STRESS input file
- `step_rotation_deg` (default 0.5°) – Rotation step in degree
- `range_rotation_deg` (default 90°) – Rotation (+/-) range in degree, for 90° total range 180°
- `image_center` (default (4.5mm, 4.5mm)) in mm
- `radius` (default 0.5mm) in mm for catheter
- `n_points` (default 20) number of points for catheter, more points stronger influence of image center
- `write_obj` (default true)
- `rest_output_path` (default "output/rest")
- `stress_output_path` (default "output/stress")
- `interpolation_steps` (default 28)
- `bruteforce` (default false)
- `sample_size` (default 200) number of points to downsample to

CSV format:

```
Frame Index, X-coord (mm), Y-coord (mm), Z-coord (mm)
185, 5.32, 2.37, 0.0
...
```

Returns:

A `PyGeometryPair` for rest, stress. A 4-tuple of `Vec<id, matched_to, rel_rot_deg, total_rot_deg, tx, ty, centroid_x, centroid_y>` for (diastole logs, systole logs, diastole stress logs, syst

Example:

```
import multimodars as mm
rest, stress, _ = mm.from_file_doublepair(
    "data/ivus_rest", "data/ivus_stress"
)
```

**multimodars.from_file_singlepair**(*input_path, step_rotation_deg=0.5, range_rotation_deg=90.0, image_center=Ellipsis, radius=0.5, n_points=20, write_obj=True, output_path='output/singlepair', interpolation_steps=28, bruteforce=False, sample_size=500*)

Processes two geometries (rest and stress) in parallel from an input CSV, returning a single `PyGeometryPair` for the chosen phase.

```
Rest/Stress pipeline:
    diastole
       |
       ▼
    systole
```

**Parameters:**
- **input_path** (-)
- **degree** (- *range_rotation_deg (default 90°) – Rotation (+/-) range in*)
- **degree**
- **180°** (*for 90° total range*)
- **(4.5mm** (- *image_center (default*)
- **4.5mm))** (*Center coordinates (x, y).*)
- **radius.** (- *radius (default 0.5mm) Processing*)
- **points.** (- *n_points (default 20) Number of boundary*)
- **true)** (- *write_obj (default*)
- **output_path** (-)
- **steps.** (- *interpolation_steps (default 28) Number of interpolation*)
- **false)** (- *bruteforce (default*)
- **to** (- *sample_size (default 200) number of points to downsample*)
- **Format** (*CSV*)
- **----------**
- **is** (*The CSV must have no header. Each row*)
- **code-block:** (..) – text: 185, 5.32, 2.37, 0.0 ...

**Returns:**
- 
- A single `PyGeometryPair` for (rest or stress) geometry.
- A 2-tuple of `Vec<id, matched_to, rel_rot_deg, total_rot_deg, tx, ty, centroid_x, centroid_y>`
- *for (diastole logs, systole logs).*

**Raises:** **RuntimeError` if the Rust pipeline fails** –

**Example**

```
import multimodars as mm
pair, _ = mm.from_file_singlepair(
    "data/ivus_rest.csv",
    "output/rest"
)
```

This is a thin Python wrapper around the Rust implementation.

**multimodars.from_file_single**(*input_path, step_rotation_deg=0.5, range_rotation_deg=90.0, diastole=True, image_center=Ellipsis, radius=0.5, n_points=20, write_obj=True, output_path='output/single', bruteforce=False, sample_size=500*)

Processes a single geometry (either diastole or systole) from an IVUS CSV file.

```
Rest/Stress pipeline (choose phase via `diastole` flag):
  e.g. diastole
```

**Parameters:**
- **input_path** (-)
- **degree** (- *range_rotation_deg (default 90°) – Rotation (+/-) range in*)
- **degree**
- **180°** (*for 90° total range*)
- **true)** (- *write_obj (default)*
- **(4.5mm** (- *image_center (default)*
- **4.5mm))** (*(x, y) center for processing.*)
- **(default** (- *output_path*)
- **(default**
- **true)**
- **(default**
- **false)** (- *bruteforce (default)*
- **to** (- *sample_size (default 200) number of points to downsample*)

**Returns:**
- 
- A `PyGeometry` containing the processed contour for the chosen phase.
- A `Vec<id, matched_to, rel_rot_deg, total_rot_deg, tx, ty, centroid_x, centroid_y>`.

**Raises:** **RuntimeError` if the underlying Rust pipeline fails** –

**Example**

```python
import multimodars as mm
geom, _ = mm.from_file_single(
    "data/ivus.csv",
    steps_best_rotation=0.5,
    range_rotation_rad=90,
    output_path="out/single",
    diastole=False
)
```

**multimodars.from_array_full**(*rest_geometry_dia, rest_geometry_sys, stress_geometry_dia, stress_geometry_sys, step_rotation_deg=0.5, range_rotation_deg=90.0, image_center=Ellipsis, radius=0.5, n_points=20, write_obj=True, rest_output_path='output/rest', stress_output_path='output/stress', diastole_output_path='output/diastole', systole_output_path='output/systole', interpolation_steps=28, bruteforce=False, sample_size=200*)

Process four existing `PyGeometry` objects (rest-dia, rest-sys, stress-dia, stress-sys) in parallel, aligning and interpolating between phases.

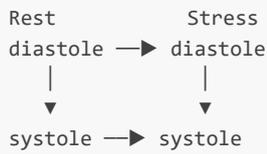

```
Rest          Stress
diastole ───▶ diastole
   │             │
   ▼             ▼
systole ───▶ systole
```

| | |
|---|---|
| **Parameters:** | • **rest_geometry_dia** (-) |
| | • **rest_geometry_sys** (-) |
| | • **stress_geometry_dia** (-) |
| | • **stress_geometry_sys** (-) |
| | • **degree** (- *range_rotation_deg (default 90°) – Rotation (+/-) range in*) |
| | • **degree** |
| | • **180°** (*for 90° total range*) |
| | • **true)** (- *write_obj (default)* |
| | • **(default** (- *interpolation_steps*) |
| | • **(default** |
| | • **(default** |
| | • **(default** |
| | • **(default** |
| | • **false)** (- *bruteforce (default)* |
| | • **to** (- *sample_size (default 200) number of points to downsample*) |
| **Returns:** | • |
| | • A `PyGeometryPair` for rest, stress, diastole, systole. |
| | • A 4-tuple of `Vec<id, matched_to, rel_rot_deg, total_rot_deg, tx, ty, centroid_x, centroid_y>` |
| | • *for (diastole logs, systole logs, diastole stress logs, systole stress logs).* |
| **Raises:** | **RuntimeError` if any Rust-side processing fails** – |

**Example**

```python
import multimodars as mm
# Assume you have four PyGeometry objects from earlier:
rest, stress, dia, sys, _ = mm.from_array_full(
    rest_dia, rest_sys, stress_dia, stress_sys,
    steps_best_rotation=0.1,
    interpolation_steps=28,
    rest_output_path="out/rest",
    stress_output_path="out/stress"
)
rest_pair, stress_pair, dia_pair, sys_pair = full
```

**multimodars.from_array_doublepair**(*rest_geometry_dia, rest_geometry_sys, stress_geometry_dia, stress_geometry_sys, step_rotation_deg=0.5, range_rotation_deg=90.0, image_center=Ellipsis, radius=0.5, n_points=20, write_obj=True, rest_output_path='output/rest', stress_output_path='output/stress', interpolation_steps=28, bruteforce=False, sample_size=500*)

Processes two geometries in parallel.

Pipeline:

```
Rest:                    Stress:
diastole                 diastole
    |                        |
    v                        v
systole                  systole
```

**Parameters:**
- **rest_geometry_dia** (-)
- **rest_geometry_sys** (-)
- **stress_geometry_dia** (-)
- **stress_geometry_sys** (-)
- **degree.** (- *step_rotation_deg (default 0.5°) – Rotation step in*)
- **degree** (- *range_rotation_deg (default 90°) – Rotation (+/-) range in*)
- **180°.** (*for 90° total range*)
- **true)** (- *write_obj (default)*
- **(default** (- *interpolation_steps)*
- **(default**
- **(default**
- **false)** (- *bruteforce (default)*
- **to** (- *sample_size (default 200) number of points to downsample*)

**Returns:**
- 
- A tuple `(rest_pair, stress_pair)` of type `(PyGeometryPair, PyGeometryPair)` ,
- *containing the interpolated diastole/systole geometries for REST and STRESS.*
- A 4-tuple of `Vec<id, matched_to, rel_rot_deg, total_rot_deg, tx, ty, centroid_x, centroid_y>`
- *for (diastole logs, systole logs, diastole stress logs, systole stress logs).*

**Raises:**    **RuntimeError`** **if any Rust-side processing fails** –

**Example**

```python
import multimodars as mm
rest_pair, stress_pair, _ = mm.from_array_doublepair(
    rest_dia, rest_sys,
    stress_dia, stress_sys,
    steps_best_rotation=0.2,
    interpolation_steps=32,
    rest_output_path="out/rest",
    stress_output_path="out/stress"
)
```

**multimodars.from_array_singlepair**(*geometry_dia, geometry_sys, step_rotation_deg=0.5, range_rotation_deg=90.0, image_center=Ellipsis, radius=0.5, n_points=20, write_obj=True, output_path='output/singlepair', interpolation_steps=28, bruteforce=False, sample_size=500*)

Interpolate between two existing `PyGeometry` objects (diastole and systole) and return a `PyGeometryPair` containing both phases.

```
Geometry pipeline:
  diastole ──▶ systole
```

**Parameters:**
- **geometry_dia** (-)
- **geometry_sys** (-)
- **degree** (- *range_rotation_deg (default 90°) – Rotation (+/-) range in*)
- **degree**
- **180°** (*for 90° total range*)
- **true)** (- *write_obj (default)*)
- **output_path** (-)
- **(default** (- *interpolation_steps*)
- **false)** (- *bruteforce (default)*)
- **to** (- *sample_size (default 200) number of points to downsample*)

**Returns:**
-
- A `PyGeometryPair` tuple containing the diastole and systole geometries with interpolation applied.
- A 2-tuple of `Vec<id, matched_to, rel_rot_deg, total_rot_deg, tx, ty, centroid_x, centroid_y>`
- *for (diastole logs, systole logs).*

**Raises:**    **RuntimeError` if the underlying Rust function fails** –

**Example**

```
import multimodars as mm
pair, _ = mm.from_array_singlepair(
    rest_dia, rest_sys,
    output_path="out/single",
    steps_best_rotation=0.1,
    interpolation_steps=30
)
```

**multimodars.geometry_from_array**(*geometry, step_rotation_deg=0.5, range_rotation_deg=90.0, image_center=Ellipsis, radius=0.5, n_points=20, label='None', records=None, delta=0.0, max_rounds=5, diastole=True, sort=True, write_obj=False, output_path='output/single', bruteforce=False, sample_size=500*)

Process an existing `PyGeometry` by optionally reordering, aligning, and refining its contours, walls, and catheter data based on various criteria.

This wraps the internal Rust function `geometry_from_array_rs`, which: 1. Builds catheter contours (if `n_points > 0` ), 2. Optionally reorders contours using provided `records` and z-coordinate sorting, 3. Aligns frames and refines the contour ordering via dynamic programming or 2-opt, 4. Smooths the final geometry, 5. Optionally writes OBJ meshes.

**Parameters:**
- **geometry** (-)
- **degree** (- *range_rotation_deg (default 90°) – Rotation (+/-) range in*)
- **degree**
- **180°** (*for 90° total range*)
- **(default** (- *output_path*)
- **(default**
- **(default**
- **(default**
- **(default**
- **(default**
- **(default**

- **(default
- **(default
- **(default
- **(default
- **false)** (- *bruteforce (default)*
- **to** (- *sample_size (default 200) number of points to downsample)*

**Returns:**

- 
- A new `PyGeometry` instance containing reordered, aligned, and smoothed contours.
- A `Vec<id, matched_to, rel_rot_deg, total_rot_deg, tx, ty, centroid_x, centroid_y>`.

**Raises:** **RuntimeError` if any Rust-side processing step fails** –

## Example

```python
import multimodars as mm
# Suppose ``geo`` is an existing PyGeometry from earlier processing:
refined, _ = mm.geometry_from_array(
    geo,
    steps_best_rotation=200,
    range_rotation_rad=1.0,
    records=my_records,
    delta=0.2,
    max_rounds=3,
    sort=True,
    write_obj=True,
    output_path="out/mesh"
)
```

# Centerline Functions

**multimodars.align_three_point**(*centerline, geometry_pair, aortic_ref_pt, upper_ref_pt, lower_ref_pt, angle_step_deg=1.0, write=False, interpolation_steps=28, output_dir='output/aligned', case_name='None'*)

Creates centerline-aligned meshes for diastolic and systolic geometries based on three reference points (aorta, upper section, lower section). Only works for elliptic vessels e.g. coronary artery anomalies.

| Parameters: | • **centerline** – PyCenterline object |
| --- | --- |
| | • **geometry_pair** – PyGeometryPair object |
| | • **aortic_ref_pt** – Reference point for aortic position |
| | • **upper_ref_pt** – Upper reference point |
| | • **lower_ref_pt** – Lower reference point |
| | • **state** – Physiological state ("rest" or "stress") |
| | • **input_dir** – Input directory for raw geometries |
| | • **output_dir** – Output directory for aligned meshes |
| | • **interpolation_steps** – Number of interpolation steps |
| Returns: | PyGeometryPair, PyCenterline (resampled) |

**Example**

```
>>> import multimodars as mm
>>> dia, sys = mm.align_three_point(
...     centerline,
...     geometry_pair,
...     (12.2605, -201.3643, 1751.0554),
...     (11.7567, -202.1920, 1754.7975),
...     (15.6605, -202.1920, 1749.9655),
... )
```

**multimodars.align_manual**(*centerline, geometry_pair, rotation_angle, start_point, write=False, interpolation_steps=28, output_dir='output/aligned', case_name='None'*)

Creates centerline-aligned meshes for diastolic and systolic geometries based on three reference points (aorta, upper section, lower section). Only works for elliptic vessels e.g. coronary artery anomalies.

| Parameters: | • **centerline** – PyCenterline object |
| --- | --- |
| | • **geometry_pair** – PyGeometryPair object |
| | • **aortic_ref_pt** – Reference point for aortic position |
| | • **upper_ref_pt** – Upper reference point |
| | • **lower_ref_pt** – Lower reference point |
| | • **state** – Physiological state ("rest" or "stress") |
| | • **input_dir** – Input directory for raw geometries |
| | • **output_dir** – Output directory for aligned meshes |
| | • **interpolation_steps** – Number of interpolation steps |
| Returns: | PyGeometryPair |

latest

## Example

```
>>> import multimodars as mm
>>> dia, sys = mm.centerline_align(
...     "path/to/centerline.csv",
...     (1.0, 2.0, 3.0),
...     (4.0, 5.0, 6.0),
...     (7.0, 8.0, 9.0),
...     "rest"
... )
```

# Utility Functions

---

**multimodars.create_catheter_geometry**(*geometry, image_center=Ellipsis, radius=0.5, n_points=20*)

Generate catheter contours and return a new PyGeometry with them filled in.

This function takes an existing PyGeometry, extracts all its contour points, computes catheter contours around those points, and returns a new PyGeometry with the catheter field populated.

| | |
|---|---|
| **Parameters:** | • **geometry** (-) |
| | • **image_center** (-) |
| | • **radius** (-) |
| | • **n_points** (-) |
| **Return type:** | A new `PyGeometry` with the same data but *catheter* field filled. |

---

**multimodars.to_obj**(*geometry, output_path, contours=True, walls=True, catheter=True, filename_contours='contours.obj', material_contours='contours.mtl', filename_catheter='catheter.obj', material_catheter='catheter.mtl', filename_walls='walls.obj', material_walls='walls.mtl'*)

Convert a `PyGeometry` object into one or more OBJ files and write them to disk.

This function takes a Python-exposed geometry ( `PyGeometry` ), converts it into the corresponding Rust geometry, and writes the specified components (contours, walls, catheter) as OBJ meshes without UV coordinates. Each component is written to its own file, with a corresponding MTL material file.

# Arguments

- `geometry` - Input `PyGeometry` instance containing the mesh data.
- `output_path` - Directory path where the OBJ and MTL files will be written.
- `contours` - Whether to export the contour mesh (default: `true` ).
- `walls` - Whether to export the wall mesh (default: `true` ).
- `catheter` - Whether to export the catheter mesh (default: `true` ).
- `filename_contours` - Filename for the contour OBJ (default: "contours.obj").
- `material_contours` - Filename for the contour MTL (default: "contours.mtl").
- `filename_catheter` - Filename for the catheter OBJ (default: "catheter.obj").
- `material_catheter` - Filename for the catheter MTL (default: "catheter.mtl").
- `filename_walls` - Filename for the walls OBJ (default: "walls.obj").
- `material_walls` - Filename for the walls MTL (default: "walls.mtl").

# Errors

Returns a *PyRuntimeError* if any of the underlying file writes fail.

latest

# Converters

---

**multimodars._converters.numpy_to_centerline**(*arr, aortic=False*)    [source]

Build a PyCenterline from a numpy array of shape (N,3), where each row is (x, y, z).

| | |
|---|---|
| **Parameters:** | • **arr** ( `ndarray` ) – np.ndarray of shape (N,3) |
| | • **aortic** ( `bool` ) – whether to mark each point as aortic |
| **Return type:** | `PyCenterline` |
| **Returns:** | PyCenterline |

---

**multimodars._converters.numpy_to_geometry**(*contours_arr, catheters_arr, walls_arr, reference_arr*)    [source]

Build a PyGeometry from four (M, 4) NumPy arrays or structured arrays, one per layer, grouping by frame_index.

Each row in the `*_arr` is [frame_index, x, y, z].

| | |
|---|---|
| **Return type:** | `PyGeometry` |
| **Parameters:** | • **contours_arr** (*ndarray*) |
| | • **catheters_arr** (*ndarray*) |
| | • **walls_arr** (*ndarray*) |
| | • **reference_arr** (*ndarray*) |

**Returns a PyGeometry containing:**

- contours: list of PyContour (one per frame in contours_arr)
- catheters: list of PyContour (one per frame in catheters_arr)
- walls: list of PyContour (one per frame in walls_arr)
- reference: single PyContourPoint from reference_arr[0]

---

**multimodars._converters.to_array**(*generic*)    [source]

Convert various multimodars Py* objects into numpy array(s) or dictionaries of arrays.

| | |
|---|---|
| **Parameters:** | **generic** (*PyContour*, *PyCenterline*, *PyGeometry*, *or PyGeometryPair*) – The object to be converted to numpy representation. |
| **Return type:** | `Union` [ `ndarray` , `dict` , `Tuple` [ `dict` , `dict` ]] |
| **Returns:** | |

- *np.ndarray* – For PyContour or PyCenterline: A 2D array of shape (N, 4), where each row is (frame_index, x, y, z).
- *dict[str, np.ndarray]* – For PyGeometry: A dictionary with keys ["contours", "catheters", "walls", "reference"], each containing a 2D array of shape (M, 4), where M is the number of points in that layer. "reference" is a (1, 4) array or (0, 4) if missing.
- *Tuple[dict[str, np.ndarray], dict[str, np.ndarray]]* – For PyGeom... tuple of two dictionaries (one for diastolic, one for systolic), ... same format as returned for a single PyGeometry.

**Raises:** **TypeError** – If the input type is not one of the supported multimodars types.

## Wrappers

multimodars._wrappers.**from_array**(*mode*, ***kwargs*) [source]

Unified entry for all array-based pipelines.

**Parameters:**
- **mode** (*{"full", "doublepair", "singlepair", "single"}*) – Which array-based pipeline to run.
- ****kwargs** (*dict*) – Keyword-only arguments required vary by mode (see below).
- **Modes** (*Supported*)
- ---------------
- **"full"** (*-*) – **from_array_full(rest_dia, rest_sys, stress_dia, stress_sys,**

    step_rotation_deg, range_rotation_deg, interpolation_steps, rest_output_path, stress_output_path, diastole_output_path, systole_output_path, image_center, radius, n_points)

- **"doublepair"** (*-*) – **from_array_doublepair(rest_dia, rest_sys, stress_dia, stress_sys,**

    step_rotation_deg, range_rotation_deg, interpolation_steps, rest_output_path, stress_output_path, image_center, radius, n_points)

- **"singlepair"** (*-*) – **from_array_singlepair(rest_dia, rest_sys, output_path,**

    step_rotation_deg, range_rotation_deg, interpolation_steps, image_center, radius, n_points)

- **"single"** (*-*) – **geometry_from_array(contours, walls, reference_point,**

    steps, range, image_center, radius, n_points, label, records, delta, max_rounds, diastole, sort, write_obj, output_path)

**Return type:** Union [ Tuple [ PyGeometryPair , PyGeometryPair , PyGeometryPair , PyGeometryPair , list , list , list , list ], Tuple [ PyGeometryPair , PyGeometryPair , list , list , list , list ], Tuple [ PyGeometryPair , PyGeometryPair , list , list ], Tuple [ PyGeometry , list ]]

**Returns:**
- Depends on *mode*
- - *"full"* –

    **Tuple[PyGeometryPair, PyGeometryPair, PyGeometryPair, PyGeometryPair,**

        list, list, list, list]

- - *"doublepair"* – Tuple[PyGeometryPair, PyGeometryPair, list, list, list, list]
- - *"singlepair"* – Tuple[PyGeometryPair, PyGeometryPair, list, list]
- - *"single"* – Tuple[PyGeometry, list]

**Raises:** **ValueError** – If an unsupported *mode* is passed.

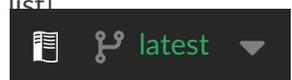

**multimodars._wrappers.from_file**(*mode, **kwargs*)    [source]

A unified entrypoint for all *from_file_** variants.

| Parameters: | • **mode** (*{"full","doublepair","singlepair","single"}*) – Selects which low-level function is called. |
|---|---|

- **"full" → needs rest_input_path, stress_input_path,**

  rest_output_path, stress_output_path, diastole_output_path, systole_output_path, step_rotation_deg, range_rotation_deg, interpolation_steps, image_center, radius, n_points

- **"doublepair" → needs rest_input_path, stress_input_path,**

  rest_output_path, stress_output_path, step_rotation_deg, range_rotation_deg, interpolation_steps, image_center, radius, n_points

- **"singlepair" → needs input_path, output_path,**

  step_rotation_deg, range_rotation_deg, interpolation_steps, image_center, radius, n_points

- **"single" → needs input_path, output_path,**

  step_rotation_deg, range_rotation_deg, diastole, image_center, radius, n_points

- **\*\*kwargs** (*dict*) – Keyword arguments required depend on *mode* (see above).

**Return type:** Union [ Tuple [ PyGeometryPair , PyGeometryPair , PyGeometryPair , PyGeometryPair , list , list , list , list ], Tuple [ PyGeometryPair , PyGeometryPair , list , list , list , list ], Tuple [ PyGeometryPair , PyGeometryPair , list , list ], Tuple [ PyGeometryPair , list ]]

**Returns:**

- *Union[* – # full now returns 4 geometry pairs *and* 4 log-lists Tuple[ PyGeometryPair, PyGeometryPair, PyGeometryPair, PyGeometryPair, list, list, list, list, ], # doublepair returns 2 geom + 4 log-lists Tuple[ PyGeometryPair, PyGeometryPair, list, list, list, list, ], # singlepair returns 1 geom-pair + 2 log-lists Tuple[ PyGeometryPair, PyGeometryPair, list,list, ], # single returns 1 geom + 1 log-list Tuple[ PyGeometryPair, list, ],
- *]* – The exact return shape depends on *mode*.

**Raises:** **ValueError** – If an unsupported *mode* is passed.

---

**multimodars._wrappers.to_centerline**(*mode, **kwargs*)    [source]

Unified entry for all to_centerline pipelines.

# Supported modes

::

- **"three_pt" → align_three_point(centerline, geometry_pair, aortic_ref_pt, upper_ref_pt,**

  lower_ref_pt, angle_step, write, interpolation_steps, output_dir, case_na

- **"manual" → align_manual(centerline, geometry_pair, rotation_angle, start_point,**

write, interpolation_steps, output_dir, case_name)

| | |
|---|---|
| **type mode:** | `Literal [ 'three_pt' , 'manual' ]` |
| **param mode:** | Which array-based pipeline to run (see "Supported modes" above). |
| **type mode:** | {"three_pt","manual"} |
| **type **kwargs:** | `Any` |
| **param **kwargs:** | Keyword-only arguments required vary by mode (see above). |
| **type **kwargs:** | dict |
| **returns:** | Depends on *mode*. |
| **rtype:** | PyGeometryPair |
| **raises ValueError:** | If an unsupported *mode* is passed. |
| **Parameters:** | • **mode** (*Literal['three_pt', 'manual']*) |
| | • **kwargs** (*Any*) |
| **Return type:** | *Tuple*[PyGeometryPair, PyCenterline] |